\documentclass{article}
\pdfoutput=1
\usepackage{microtype}
\usepackage{graphicx}
\usepackage{subfigure}
\usepackage{booktabs} %

\usepackage{hyperref}

\usepackage[accepted]{icml2019}

\icmltitlerunning{A Theory of Regularized MDPs}

\usepackage{amsmath}
\usepackage{amsthm}
\usepackage{dsfont}
\usepackage{amssymb}
\usepackage{amscd}
\usepackage{mathtools}
\mathtoolsset{showonlyrefs=true}
\usepackage{enumerate}

\newtheorem{definition}{Definition}
\newtheorem{theorem}{Theorem}
\newtheorem{prop}{Proposition}

\newtheorem{lemma}{Lemma}
\newtheorem{cor}{Corollary}

\newcommand{\argmax}{\operatorname*{argmax}} %

\newcommand{\kl}{\operatorname*{KL}}

\newcommand\myeq[2]{\stackrel{\mathclap{(#1)}}{#2}}

\newcommand{\E}{\mathbb{E}}

\newcommand{\g}{\mathcal{G}}

\newcommand{\R}{\mathbb{R}}

\newcommand{\s}{\mathcal{S}}
\newcommand{\A}{\mathcal{A}}

\newcommand{\gc}{\mathcal{G}}

\newcommand{\un}{\mathbf{1}}

\begin{document}

\twocolumn[
\icmltitle{A Theory of Regularized Markov Decision Processes}

\icmlsetsymbol{equal}{*}

\begin{icmlauthorlist}
\icmlauthor{Matthieu Geist}{google}
\icmlauthor{Bruno Scherrer}{inria} %
\icmlauthor{Olivier Pietquin}{google} %
\end{icmlauthorlist}

\icmlaffiliation{google}{Google Research, Brain Team.}
\icmlaffiliation{inria}{Universit\'e de Lorraine, CNRS, Inria, IECL, F-54000 Nancy, France}

\icmlcorrespondingauthor{Matthieu Geist}{mfgeist@google.com}

\icmlkeywords{Reinforcement Learning, Markov Decision Processes, Regularization, Mirror Descent, Bregman Divergence}

\vskip 0.3in
]

\printAffiliationsAndNotice{}  %

\begin{abstract}
Many recent successful (deep) reinforcement learning algorithms make use of regularization, generally based on entropy or Kullback-Leibler divergence. We propose a general theory of regularized Markov Decision Processes that generalizes these approaches in two directions: we consider a larger class of regularizers, and we consider the general modified policy iteration approach, encompassing both policy iteration and value iteration. The core building blocks of this theory are a notion of regularized Bellman operator and the Legendre-Fenchel transform, a classical tool of convex optimization. This approach allows for error propagation analyses of general algorithmic schemes of which (possibly variants of) classical algorithms such as Trust Region Policy Optimization, Soft Q-learning, Stochastic Actor Critic or Dynamic Policy Programming are special cases. This also draws connections to proximal convex optimization, especially to Mirror Descent.%
\end{abstract}

\section{Introduction}

Many reinforcement learning algorithms make use of some kind of entropy regularization, with various motivations, such as improved exploration and robustness. Trust Region Policy Optimization (TRPO)~\cite{schulman2015trust} is a policy iteration scheme where the greedy step is penalized with a Kullback-Leibler (KL) penalty between two consecutive policies. Dynamic Policy Programming (DPP)~\cite{azar2012dynamic} is a reparametrization of a value iteration scheme regularized by a KL penalty between consecutive policies. Soft Q-learning, eg.~\cite{fox2015taming,schulman2017equivalence,haarnoja2017reinforcement}, uses a Shannon entropy regularization in a value iteration scheme, while Soft Actor Critic (SAC)~\cite{haarnoja2018soft} uses it in a policy iteration scheme. Value iteration has also been combined with a Tsallis entropy~\cite{lee2017sparse}, with the motivation of having a sparse regularized greedy policy. Other approaches are based on a notion of temporal consistency equation, somehow extending the notion of Bellman residual to the regularized case~\cite{nachum2017bridging,dai2018sbeed,nachum2018path}, or on  policy gradient~\cite{williams1992simple,mnih2016asynchronous}.

This non-exhaustive set of algorithms share the idea of using regularization, but they are derived from sometimes different principles, consider each time a specific regularization, and have ad-hoc analysis, if any. Here, we propose a general theory of regularized Markov Decision Processes (MDPs). To do so, a key observation is that (approximate) dynamic programming, or (A)DP, can be derived solely from the core definition of the Bellman evaluation operator. The framework we propose is built upon a regularized Bellman operator, and on an associated Legendre-Fenchel transform. We study the theoretical properties of these regularized MDPs and of the related regularized ADP schemes. This generalizes many existing theoretical results and provides new ones. Notably, it allows for an error propagation analysis
for many of the aforementioned algorithms. This framework also draws connections to convex optimization, especially to Mirror Descent (MD).

A unified view of entropy-regularized MDPs has already been proposed by~\citet{neu2017unified}. They focus on regularized DP through linear programming for the average reward case. Our contribution is complementary to this work (different MDP setting, we do not regularize the same quantity, we do not consider the same DP approach). Our use of the Legendre-Fenchel transform is inspired by~\citet{mensch2018differentiable}, who consider smoothed finite horizon DP in directed acyclic graphs. Our contribution is also complementary to this work, that does not allow recovering aforementioned algorithms nor analyzing them. 
After a brief background, we introduce regularized MDPs and various related algorithmic schemes based on approximate modified policy iteration~\cite{scherrer2015approximate}, as well as their analysis. All proofs are provided in the appendix.

\section{Background}

In this section, we provide the necessary background for building the proposed regularized MDPs. We write $\Delta_X$ the set of probability distributions over a finite set $X$ and $Y^X$ the set of applications from $X$ to the set $Y$.
All vectors are column vectors, except distributions, for left multiplication. 
We  write $\langle\cdot,\cdot\rangle$ the  dot product and $\|\cdot\|_p$ the $\ell_p$-norm.

\subsection{Unregularized MDPs}

An MDP is a tuple $\{\s,\A,P, r, \gamma\}$ with $\s$ the finite\footnote{We assume a finite space for simplicity of exposition, our results extend to more general cases.} state space, $\A$ the finite action space, $P\in\Delta_\s^{\s\times \A}$ the Markovian transition kernel ($P(s'|s,a)$ denotes the probability of transiting to $s'$ when action $a$ is applied in state $s$), $r\in \R^{\s\times \A}$ the reward function and $\gamma\in(0,1)$ the discount factor.

A policy $\pi \in \Delta_\A^\s$ associates to each state a distribution over actions. The associated Bellman operator is defined as, for any function $v\in\R^\s$, 
\begin{equation}
    \forall s \in \s, \, [T_\pi v](s) = \E_{a\sim\pi(.|s)}\left[r(s,a) + \gamma \E_{s'|s,a} [v(s')]\right].
\end{equation}
This operator is a $\gamma$-contraction in supremum norm and its unique fixed-point is the value function $v_\pi$. With $r_\pi(s) =  \E_{a\sim\pi(.|s)}[r(s,a)]$ and $P_\pi(s'|s) = \E_{a\sim\pi(.|s)}[P(s'|s,a)])$, the operator can be written as $T_\pi v = r_\pi + \gamma P_\pi v$. For any function $v\in\R^\s$, we associate the function $q\in\R^{\s\times\A}$,%
\begin{equation}
    q(s,a) = r(s,a) + \gamma \E_{s'|s,a} [v(s')].
\end{equation}
Thus, the Bellman operator can also be written as $[T_\pi v](s) = \langle\pi(\cdot|s), q(s,\cdot)\rangle = \langle\pi_s, q_s\rangle$. With a slight abuse of notation, we will write $T_\pi v = \langle \pi, q\rangle = (\langle\pi_s, q_s\rangle)_{s\in\s}$.

From this evaluation operator, one can define the Bellman optimality operator as, for any $v\in\R^\s$,
\begin{equation}
    T_* v = \max_\pi T_\pi v.
\end{equation}
This operator is also a $\gamma$-contraction in supremum norm, and its fixed point is the optimal value function $v_*$.
From the same operator, one can also define the notion of a policy being greedy respectively to a function $v\in\R^\s$:
\begin{equation}
    \pi'\in\g(v) \Leftrightarrow T_* v = T_{\pi'} v \Leftrightarrow \pi' \in \argmax_{\pi} T_\pi v.
\end{equation}
Given this, we could derive value iteration, policy iteration, modified policy iteration, and so on. Basically, we can do all these things from the core definition of the Bellman evaluation operator. We'll do so from a notion of regularized Bellman evaluation operator.

\subsection{Legendre-Fenchel transform}

Let $\Omega:\Delta_\A\rightarrow\R$ be a strongly convex function. The Legendre-Fenchel transform (or convex conjugate) of $\Omega$ is $\Omega^*:\R^\A\rightarrow\R$, defined as
\begin{equation}
    \forall q_s\in\R^\A,\, \Omega^*(q_s) = \max_{\pi_s\in\Delta_\A}\langle \pi_s, q_s\rangle - \Omega(\pi_s).
\end{equation}
We'll make use of the following properties~\cite{hiriart2012fundamentals,mensch2018differentiable}.

\begin{prop}
\label{prop:convex-conjugate}
    Let $\Omega$ be strongly convex, we have the following properties.
    \begin{enumerate}[i]
    \item Unique maximizing argument: $\nabla \Omega^*$ is Lipschitz and satisfies
        $\nabla \Omega^*(q_s) = \argmax_{\pi_s\in\Delta_\A} \langle \pi_s, q_s\rangle - \Omega(\pi_s)$.
    \item Boundedness: if there are constants $L_\Omega$ and $U_\Omega$ such that for all $\pi_s\in\Delta_\A$, we have $L_\Omega \leq \Omega(\pi_s) \leq U_\Omega$, then
        $\max_{a\in\A} q_s(a) - U_\Omega \leq \Omega^*(q_s) \leq \max_{a\in\A} q_s(a) - L_\Omega$.
    \item Distributivity: for any $c\in\R$ (and $\un$ the vector of ones), we have
        $\Omega^*(q_s + c \un) = \Omega^*(q_s) + c$. 
    \item Monotonicity:
        $q_{s,1}\leq q_{s,2} \Rightarrow \Omega^*(q_{s,1})\leq \Omega^*(q_{s,2})$.
\end{enumerate} 
\end{prop}

A classical example is the negative entropy $\Omega(\pi_s) = \sum_a \pi_s(a)\ln\pi_s(a)$. Its convex conjugate is the smoothed maximum $\Omega^*(q_s)= \ln \sum_a \exp q_s(a)$ and the unique maximizing argument is the usual softmax $\nabla\Omega^*(q_s) = \frac{\exp q_s(a)}{\sum_b \exp q_s(b)}$. For a positive regularizer, one can consider $\Omega(\pi_s) = \sum_a \pi_s(a)\ln\pi_s(a)+\ln|\A|$, that is the KL divergence between $\pi_s$ and a uniform distribution. Its convex conjugate is $\Omega^*(q_s)= \ln \sum_a \frac{1}{|\A|} \exp q_s(a)$, that is the Mellowmax operator~\cite{asadi2016alternative}. The maximizing argument is still the softmax. Another less usual example is the negative Tsallis entropy~\cite{lee2017sparse}, $\Omega(\pi_s) = \frac{1}{2}(\|\pi_s\|_2^2 - 1)$. The analytic convex conjugate is more involved, but it leads to the sparsemax as the maximizing argument~\cite{martins2016softmax}.

\section{Regularized MDPs}
\label{sec:reg-mpds}

The core idea of our contribution is to regularize the Bellman evaluation operator. Recall that $[T_\pi v](s) = \langle\pi_s,q_s\rangle$. A natural idea is to replace it by $[T_{\pi,\Omega} v](s) = \langle\pi_s,q_s\rangle - \Omega(\pi_s)$. To get the related optimality operator, one has to perform state-wise maximization over $\pi_s\in\Delta_\A$, which gives the Legendre-Fenchel transform of $[T_{\pi,\Omega} v](s)$. This defines a smoothed maximum~\cite{nesterov2005smooth}. The related maximizing argument defines the notion of greedy policy.

\subsection{Regularized Bellman operators}

We now define formally these regularized Bellman operators. With a slight abuse of notation, we write $\Omega(\pi) = (\Omega(\pi_s))_{s\in\s}$ (and similarly for $\Omega^*$ and $\nabla \Omega^*$).

\begin{definition}[Regularized Bellman operators]
\label{def:reg-op}
    Let $\Omega:\Delta_\A\rightarrow \R$ be a strongly convex function. For any $v\in\R^\s$ define $q\in\R^{\s\times\A}$ as $q(s,a) = r(s,a) + \gamma \E_{s'|s,a}[v(s')]$.
    The regularized Bellman evaluation operator is defined as
    \begin{equation}
        T_{\pi,\Omega}:v\in\R^\s \rightarrow T_{\pi,\Omega}v = T_\pi v -\Omega(\pi) \in \R^\s,
    \end{equation}
    that is, state-wise, $[T_{\pi,\Omega}v ](s) = \langle \pi_s, q_s \rangle - \Omega(\pi_s)$.
    The regularized Bellman optimality operator is defined as
    \begin{equation}
        T_{*,\Omega}:v\in\R^\s \rightarrow T_{*,\Omega}v = \max_{\pi\in\Delta_\A^\s}T_{\pi,\Omega} v = \Omega^*(q) \in \R^\s,
    \end{equation}
    that is, state-wise, $[T_{*,\Omega}v](s) = \Omega^*(q_s)$.
    For any function $v\in\R^\s$, the associated unique greedy policy is defined as
    \begin{equation}
        \pi'=\g_\Omega(v) = \nabla\Omega^*(q) \Leftrightarrow T_{\pi',\Omega} v = T_{*,\Omega} v,
    \end{equation}
    that is, state-wise, $\pi'_s = \nabla \Omega^*(q_s)$.
\end{definition}

To be really useful, these operators should satisfy the same properties as the classical ones. It is indeed the case (we recall that all proofs are provided in the appendix).

\begin{prop}
\label{prop:reg-op}
    The operator $T_{\pi,\Omega}$ is affine and we have the following properties.
    \begin{enumerate}[i]
        \item Monotonicity: let $v_1,v_2\in\R^\s$ such that $v_1\geq v_2$. Then, %
        \begin{equation}
            T_{\pi,\Omega} v_1 \geq T_{\pi,\Omega} v_2 \text{ and } T_{*,\Omega} v_1 \geq T_{*,\Omega} v_2.
        \end{equation}
        \item Distributivity: for any $c\in\R$, we have that
        \begin{align}
            T_{\pi,\Omega}(v+c\un) &= T_{\pi,\Omega}v + \gamma c \un  \\ \text{and }
            T_{*,\Omega}(v+c\un) &= T_{*,\Omega}v + \gamma c \un.
        \end{align}
        \item Contraction: both operators are $\gamma$-contractions in supremum norm. For any $v_1,v_2\in\R^\s$,
        \begin{align}
            \|T_{\pi,\Omega}v_1 - T_{\pi,\Omega}v_2\|_\infty &\leq \gamma \|v_1 - v_2\|_\infty
            \\ \text{and }
            \|T_{*,\Omega}v_1 - T_{*,\Omega}v_2\|_\infty &\leq \gamma  \|v_1 - v_2\|_\infty.
        \end{align}
    \end{enumerate}
\end{prop}

\subsection{Regularized value functions}

The regularized operators being contractions, we can define regularized value functions as their unique fixed-points. Notice that from the following definitions, we could also easily derive regularized Bellman operators on $q$-functions. 

\begin{definition}[Regularized value function of policy $\pi$]
\label{def:reg-value-policy}
    Noted $v_{\pi,\Omega}$, it is defined as the unique fixed point of the operator $T_{\pi,\Omega}$: $v_{\pi,\Omega} = T_{\pi,\Omega} v_{\pi,\Omega}$. We also define the associated state-action value function $q_{\pi,\Omega}$ as
    \begin{align}
        q_{\pi,\Omega}(s,a) &= r(s,a) + \gamma \E_{s'|s,a}[v_{\pi,\Omega}(s')]
        \\ \text{with }
        v_{\pi,\Omega}(s) &= \E_{a\sim\pi(.|s)}[q_{\pi,\Omega}(s,a)] - \Omega(\pi(.|s)).
    \end{align}
\end{definition}

Thus, the regularized value function is simply the unregularized value of $\pi$ for the reward $r_\pi - \Omega(\pi)$, that is $v_{\pi,\Omega} = (I-\gamma P_\pi)^{-1}(r_\pi - \Omega(\pi))$.

\begin{definition}[Regularized optimal value function]
\label{def:reg-value-opt}
    Noted $v_{*,\Omega}$, it is the unique fixed point of the operator $T_{*,\Omega}$: $v_{*,\Omega} = T_{*,\Omega} v_{*,\Omega}$. 
    We also define the associated state-action value function $q_{*,\Omega}(s,a)$ as
    \begin{align}
        q_{*,\Omega}(s,a) &= r(s,a) + \gamma \E_{s'|s,a}[v_{*,\Omega}(s')]
        \\ \text{with }
        v_{*,\Omega}(s) &= \Omega^*(q_{*,\Omega}(s,.)).
    \end{align}
\end{definition}

The function $v_{*,\Omega}$ is indeed the optimal value function, thanks to the following result.

\begin{theorem}[Optimal regularized policy]
    \label{th:opt-reg-pol}
    The policy $ \pi_{*,\Omega} = \g_{\Omega}(v_{*,\Omega})$
    is the unique optimal regularized policy, in the sense that for all $\pi\in\Delta_\A^\s$, $v_{\pi_{*,\Omega},\Omega} = v_{*,\Omega} \geq v_{\pi,\Omega}$.
\end{theorem}

When regularizing the MDP, we change the problem at hand. The following result relates value functions in (un)regularized MDPs.

\begin{prop}
    \label{prop:bound-values-reg-unreg}
    Assume that $L_\Omega\leq\Omega\leq U_\Omega$. Let $\pi$ be any policy. We have that
    $v_{\pi} -\frac{U_\Omega}{1-\gamma}\un \leq v_{\pi,\Omega} \leq v_{\pi}  - \frac{L_\Omega}{1-\gamma}\un$
    and
    $v_{*} -\frac{U_\Omega}{1-\gamma}\un \leq v_{*,\Omega} \leq v_{*}  - \frac{L_\Omega}{1-\gamma}\un$.
\end{prop}

Regularization changes the optimal policy, the next result shows how it performs in the original MDP.

\begin{theorem}
    \label{th:reg-opt-in-original-mdp}
    Assume that $L_\Omega\leq\Omega\leq U_\Omega$. We have that
    \begin{equation}
        v_* - \frac{U_\Omega - L_\Omega}{1-\gamma} \leq v_{\pi_{*,\Omega}} \leq v_*.
    \end{equation}
\end{theorem}

\subsection{Related Works}

Some of these results already appeared in the literature, in different forms and with specific regularizers. For example, the contraction of $T_{*,\Omega}$ (Prop.~\ref{prop:reg-op}) was shown in various forms, e.g.~\citep{fox2015taming,asadi2016alternative,dai2018sbeed}, as well as the relation between (un)regularized optimal value functions (Th.~\ref{th:reg-opt-in-original-mdp}), e.g.~\citep{lee2017sparse,dai2018sbeed}. The link to Legendre-Fenchel has also been considered before, e.g.~\citep{dai2018sbeed,mensch2018differentiable,richemond2017short}.

The core contribution of Sec.~\ref{sec:reg-mpds} is the regularized Bellman operator, inspired by~\citet{nesterov2005smooth} and~\citet{mensch2018differentiable}. It allows building in a principled and general way regularized MDPs, and generalizing existing results easily. More importantly, it is the core building block of regularized (A)DP, studied in the next sections. The framework and analysis we propose next rely heavily on this formalism.

\section{Regularized Modified Policy Iteration}
\label{sec:regmpi}

Having defined the notion of regularized MDPs, we still need algorithms that solve them. As the regularized Bellman operators have the same properties as the classical ones, we can apply classical dynamic programming. Here, we consider directly the modified policy iteration approach~\citep{puterman1978modified}, that we regularize (reg-MPI for short):
\begin{equation}
    \begin{cases}
        \pi_{k+1} = \gc_\Omega(v_k)
        \\
        v_{k+1} = (T_{\pi_{k+1},\Omega})^m v_k
    \end{cases}.
    \label{eq:reg-mpi}
\end{equation}
Given an initial $v_0$, reg-MPI iteratively performs a regularized greedy step to get $\pi_{k+1}$ and a partial regularized evaluation step to get $v_{k+1}$.

With $m=1$, we retrieve a regularized value iteration algorithm, that can be simplified as $v_{k+1} = T_{*,\Omega} v_k$ (as $\pi_{k+1}$ is greedy resp. to $v_k$, we have $T_{\pi_{k+1},\Omega} v_k = T_{*,\Omega} v_k$). With $m=\infty$, we obtain a regularized policy iteration algorithm, that can be simplified as $\pi_{k+1} = \gc_\Omega(v_{\pi_k,\Omega})$ (indeed, with a slight abuse of notation, $(T_{\pi_k,\Omega})^\infty v_{k-1} = v_{\pi_k,\Omega}$).

Before studying the convergence and rate of convergence of this general algorithmic scheme (with approximation), we discuss its links to state of the art algorithms (and more generally how it can be practically instantiated).

\subsection{Related algorithms}
\label{subsec:related-alg-reg-mpi}

Most existing schemes consider the negative entropy as the regularizer. Usually, it is also more convenient to work with q-functions. First, we consider the case $m=1$. In the exact case, the regularized value iteration scheme can be written
\begin{equation}
    q_{k+1}(s,a) = r(s,a) + \gamma \E_{s'|s,a}[\Omega^*(q_k(s',\cdot))].
\end{equation}
In the entropic case, $\Omega^*(q_k(s,\cdot)) = \ln \sum_a \exp q_k(s,a)$.
In an approximate setting, the q-function can be parameterized by parameters $\theta$ (for example, the weights of a neural network), write $\bar{\theta}$ the target parameters (computed during the previous iteration) and $\hat{\E}$ the empirical expectation over sampled transitions $(s_i,a_i,r_i,s'_i)$, an iteration amounts to minimize the expected loss
\begin{align}
    J(\theta) &= \hat{\E}\left[\left(\hat{q}_i - q_\theta(s_i,a_i)\right)^2\right] \label{eq:sof-ql}
    \\
    \text{with }
    \hat{q}_i &= r_i + \gamma \Omega^*(q_{\bar{\theta}}(s'_i,\cdot)).
\end{align}
Getting a practical algorithm may require more work, for example for estimating $\Omega^*(q_{\bar{\theta}}(s'_i,\cdot))$ in the case of continuous actions~\citep{haarnoja2017reinforcement}, but this is the core principle of soft Q-learning~\citep{fox2015taming,schulman2017equivalence}. This idea has also been applied using the Tsallis entropy as the regularizer~\citep{lee2017sparse}.

Alternatively, assume that $q_k$ has been estimated. One could compute the regularized greedy policy analytically, $\pi_{k+1}(\cdot|s) = \nabla \Omega^*(q_k(s,\cdot))$. Instead of computing this for any state-action couple, one can generalize this from observed transitions to any state-action couple through a parameterized policy $\pi_w$, by minimizing the KL divergence between both distributions: 
\begin{equation}
    J(w) = \hat{\E}[\kl(\pi_w(\cdot|s_i)||\nabla \Omega^*(q_k(s_i,.)))]. \label{eq:loss-mappo}
\end{equation}
This is done in SAC~\citep{haarnoja2018soft}, with an entropic regularizer (and thus $\nabla \Omega^*(q_k(s,.)) = \frac{\exp q_k(s,\cdot)}{\sum_a\exp q_k(s,a)}$). This is also done in Maximum A Posteriori Policy Optimization (MPO)~\citep{abdolmaleki2018maximum} with a KL regularizer (a case we discuss Sec.~\ref{sec:mdmpi}), or by~\citet{abdolmaleki2018relative} with more general ``conservative'' greedy policies.

Back to SAC, $q_k$ is estimated using a TD-like approach, by minimizing\footnote{Actually, a separate network is used to estimate the value function, but it is not critical here.} for the current policy $\pi$:
\begin{align}
    J(\theta) &= \hat{\E}[(\hat{q}_{i} - q_\theta(s_i,a_i))^2] \label{eq:loss-sac}
    \\
    \text{with }
    \hat{q}_{i} &= r_i + \gamma (\E_{a\sim\pi(\cdot|s'_i)}[q_{\bar{\theta}}(s'_i,a)] - \Omega(\pi(\cdot,s'_i)).
\end{align}
For SAC, we have $\Omega(\pi(\cdot,s)) = \E_{a\sim\pi(\cdot|s)}[ \ln \pi(a|s)]$ specifically (negative entropy). This approximate evaluation step corresponds to $m=1$, and SAC is therefore more a VI scheme than a PI scheme, as presented by~\citet{haarnoja2018soft} (the difference with soft Q-learning lying in how the greedy step is performed, implicitly or explicitly). It could be extended to the case $m>1$ in two ways. One possibility is to minimize $m$ times the expected loss~\eqref{eq:loss-sac}, updating the target parameter vector 
$\bar{\theta}$ between each optimization, but keeping the policy $\pi$ fixed. Another possibility is to replace the 1-step rollout of Eq.~\eqref{eq:loss-sac} by an $m$-step rollout (similar to classical $m$-step rollouts, up to the additional regularizations correcting the rewards). Both are equivalent in the exact case, but not in the general case.

Depending on the regularizer, $\Omega^*$ or $\nabla \Omega^*$ might not be known analytically. In this case, one can still solve the greedy step directly. Recall that the regularized greedy policy satisfies $\pi_{k+1} = \max_\pi T_{\pi,\Omega} v_k$. In an approximate setting, this amounts to maximize\footnote{One could add a state-dependant baseline to $q_k$, eg. $v_k$, this does not change the maximizer but can reduce the variance.}
\begin{equation}
    J(w) = \hat{\E}\left[\E_{a\sim\pi_w(\cdot|s_i)}[q_k(s_i,a)] - \Omega(\pi_w(\cdot|s_i)\right]. \label{eq:loss-trpo}
\end{equation}
This improvement step is used by~\citet{riedmiller2018learning} with an entropy, as well as by TRPO (up to the fact that the objective is constrained rather than regularized), with a KL regularizer (see Sec.~\ref{sec:mdmpi}).

To sum up, for any regularizer $\Omega$, with $m=1$ one can concatenate greedy and evaluation steps as in Eq.~\eqref{eq:sof-ql}, with $m\geq 1$ one can estimate the greedy policy using either Eqs.~\eqref{eq:loss-mappo} or~\eqref{eq:loss-trpo}, and estimate the q-function using Eq.~\ref{eq:loss-sac}, either performed $m$ times repeatedly or combined with $m$-step rollouts, possibly combined with off-policy correction such as importance sampling or Retrace~\citep{munos2016safe}.

\subsection{Analysis}
\label{subsec:analysis-regMPI}

We analyze the propagation of errors of the scheme depicted in Eq.~\eqref{eq:reg-mpi}, and as a consequence, its convergence and rate of convergence. To do so, we consider possible errors in both the (regularized) greedy and evaluation steps,
\begin{equation}
    \begin{cases}
        \pi_{k+1} = \gc^{\epsilon'_{k+1}}_\Omega(v_k)
        \\
        v_{k+1} = (T_{\pi_{k+1},\Omega})^m v_k + \epsilon_{k+1}
    \end{cases},
    \label{eq:reg-mpi-error}
\end{equation}
with $\pi_{k+1} =\gc^{\epsilon'_{k+1}}_\Omega(v_k)$ meaning that for any policy $\pi$, we have $T_{\pi,\Omega} v_k \leq T_{\pi_{k+1},\Omega} v_k + \epsilon'_{k+1}$. The following analysis is basically the same as the one of Approximate Modified Policy Iteration (AMPI)~\citep{scherrer2015approximate}, thanks to the results of Sec.~\ref{sec:reg-mpds} (especially Prop.~\ref{prop:reg-op}).

The distance we bound is the loss $l_{k,\Omega} = v_{*,\Omega} - v_{\pi_k,\Omega}$. The bound will involve the terms $d_0 = v_{*,\Omega} - v_0$ and $b_0 = v_0 - T_{\pi_1,\Omega} v_0$. It requires also defining the following.

\begin{definition}[$\Gamma$-matrix~\cite{scherrer2015approximate}]
    \label{def:Gamma}
    For $n\in\mathbb{N}^*$, $\mathbb{P}_n$ is the set of transition kernels defined as \textbf{1)} for any set of $n$ policies $\{\pi_1,\dots,\pi_n\}$, $\prod_{i=1}^n (\gamma P_{\pi_i}) \in \mathbb{P}_n$ and \textbf{2)} for any $\alpha\in(0,1)$ and $(P_1,P_2)\in\mathbb{P}_n\times\mathbb{P}_n$, $\alpha P_1 + (1-\alpha)P_2\in\mathbb{P}_n$. Any element of $\mathbb{P}_n$ is denoted $\Gamma^n$.
\end{definition}

We first state a point-wise bound on the loss. This is the same bound as for AMPI, generalized to regularized MDPs.

\begin{theorem}
    After $k$ iterations of scheme~\eqref{eq:reg-mpi-error}, we have %
    \begin{equation}
        l_{k,\Omega} \leq 2\sum_{i=1}^{k-1} \sum_{j=i}^\infty \Gamma^j |\epsilon_{k-i}| + \sum_{i=0}^{k-1} \sum_{j=i}^\infty \Gamma^j |\epsilon'_{k-i}| + h(k)
    \end{equation}
    with $h(k) = 2\sum_{j=k}^\infty\Gamma^j |d_0|$ or $h(k) = 2\sum_{j=k}^\infty \Gamma^j|b_0|$.
\end{theorem}

Next, we provide a bound on the weighted $\ell_p$-norm of the loss, defined for a distribution $\rho$ as $\|l_k\|^p_{p,\rho} = \rho|l_k|^p$. Again, this is the AMPI bound generalized to regularized MDPs.

\begin{cor}
    \label{cor:lp-bound-regMPI}
    Let $\rho$ and $\mu$ be distributions. Let $p$, $q$ and $q'$ such that $\frac{1}{q} + \frac{1}{q'}=1$. Define the concentrability coefficients %
    $%
        C_q^i = \frac{1-\gamma}{\gamma^i}\sum_{j=i}^\infty \gamma^j \max_{\pi_1,\dots,\pi_j}\left\|\frac{\rho P_{\pi_1} P_{\pi_2}\dots P_{\pi_j}}{\mu}\right\|_{q,\mu}
    $. %
    After $k$ iterations of scheme~\eqref{eq:reg-mpi-error}, the loss satisfies
    \begin{align}
        \|l_{k,\Omega}\|_{p,\rho} &\leq 2\sum_{i=1}^{k-1} \frac{\gamma^i}{1-\gamma} (C_q^i)^{\frac{1}{p}} \|\epsilon_{k-i}\|_{pq',\mu}
        \\
        &\quad + \sum_{i=0}^{k-1} \frac{\gamma^i}{1-\gamma} (C_q^i)^{\frac{1}{p}} \|\epsilon'_{k-i}\|_{pq',\mu} + g(k)
    \end{align}
    with $g(k) = \frac{2\gamma^k}{1-\gamma}(C_q^i)^{\frac{1}{p}}\min(\|d_0\|_{pq',\mu}, \|b_0\|_{pq',\mu})$.
\end{cor}

As this is the same bound (up to the fact that it deals with regularized MDPs) as the one of AMPI, we refer to~\citet{scherrer2015approximate} for a broad discussion about it. It is similar to other error propagation analyses in reinforcement learning, and generalizes those that could be obtained for regularized value or policy iteration. The factor $m$ does not appear in the bound. This is also discussed by~\citet{scherrer2015approximate}, but basically this depends on where the error is injected. We could derive a regularized version of Classification-based Modified Policy Iteration (CBMPI, see~\citet{scherrer2015approximate} again) and make it appear.

So, we get the same bound for reg-MPI that for unregularized AMPI, no better nor worse. This is a good thing, as it justifies considering regularized MDPs, but it does no explain the good empirical results of related algorithms.

With regularization, policies will be more stochastic than in classical approximate DP (that tends to produce deterministic policies). Such stochastic policies can induce lower concentrability coefficients.
We also hypothesize that regularizing the greedy step helps controlling the related approximation error, that is the $\|\epsilon'_{k-i}\|_{pq',\mu}$ terms. Digging this question would require instantiating more the algorithmic scheme and performing a finite sample analysis of the resulting optimization problems. We left this for future work, and rather pursue the general study of solving regularized MDPs, with varying regularizers now.

\section{Mirror Descent Modified Policy Iteration}
\label{sec:mdmpi}

Solving a regularized MDP provides a solution that differs from the one of the unregularized MDP (see Thm.~\ref{th:reg-opt-in-original-mdp}). The problem we address here is estimating the original optimal policy while solving regularized greedy steps.
Instead of considering a fixed regularizer $\Omega(\pi)$, the key idea is to penalize a divergence between the policy $\pi$ and the policy obtained at the previous iteration of an MPI scheme.
We consider more specifically the Bregman divergence generated by the strongly convex regularizer $\Omega$.

Let $\pi'$ be some given policy (typically $\pi_k$, when computing $\pi_{k+1}$), the Bregman divergence generated by $\Omega$ is
\begin{align}
    \Omega_{\pi'_s}(\pi_s) &= D_\Omega(\pi_s||\pi'_s)
    \\
    &= \Omega(\pi_s) - \Omega(\pi'_s) - \langle\nabla\Omega(\pi'_s), \pi_s - \pi'_s\rangle.
\end{align}
For example, the KL divergence is generated by the negative entropy: $\kl(\pi_s||\pi'_s) = \sum_a \pi_s(a)\ln \frac{\pi_s(a)}{\pi_s'(a)}$. With a slight abuse of notation, as before, we will write
\begin{equation}
    \Omega_{\pi'}(\pi) = D_\Omega(\pi||\pi')
    = \Omega(\pi) - \Omega(\pi') - \langle\nabla\Omega(\pi'), \pi - \pi'\rangle.
\end{equation}
This divergence is always positive, it satisfies $\Omega_{\pi'}(\pi')=0$, and it is strongly convex in $\pi$ (so Prop.~\ref{prop:convex-conjugate} applies).

We consider a reg-MPI algorithmic scheme with a Bregman divergence replacing the regularizer. For the greedy step, we simply consider $\pi_{k+1} = \gc_{\Omega_{\pi_k}}(v_k)$, that is
\begin{equation}
    \pi_{k+1} = \argmax_\pi \langle q_k, \pi\rangle - D_\Omega(\pi||\pi_k).
\end{equation}
This is similar to the update of the Mirror Descent (MD) algorithm in its proximal form~\citep{beck2003mirror}, with $-q_k$ playing the role of the gradient in MD. Therefore, we will call this approach Mirror Descent Modified Policy Iteration (MD-MPI). For the partial evaluation step, we can regularize according to the previous policy $\pi_k$, that is $v_{k+1} = (T_{\pi_{k+1},\Omega_{\pi_k}})^m v_k$, or according to the current policy $\pi_{k+1}$, that is $v_{k+1} = (T_{\pi_{k+1},\Omega_{\pi_{k+1}}})^m v_k$. As $\Omega_{\pi_{k+1}}(\pi_{k+1})=0$, this simplifies as $v_{k+1} = (T_{\pi_{k+1}})^m v_k$, that is a partial unregularized evaluation.

To sum up, we will consider two general algorithmic schemes based on a Bregman divergence, MD-MPI types~1 and~2 respectively defined as
\begin{equation}
    \begin{cases}
        \pi_{k+1} = \gc_{\Omega_{\pi_{k}}}(v_k)
        \\
        v_{k+1} = (T_{\pi_{k+1},\Omega_{\pi_{k}}})^m v_k
    \end{cases}
    ,
    \begin{cases}
        \pi_{k+1} = \gc_{\Omega_{\pi_{k}}}(v_k)
        \\
        v_{k+1} = (T_{\pi_{k+1}})^m v_k
    \end{cases}
\end{equation}
and both initialized with some $v_0$ and $\pi_0$.

\subsection{Related algorithms}
\label{subsec:related-algs-mdmpi}

To derive practical algorithms, the recipes provided in Sec.~\ref{subsec:related-alg-reg-mpi} still apply, just replacing $\Omega$ by $\Omega_{\pi_k}$. If $m=1$, greedy and evaluation steps can be concatenated (only for MD-MPI type~1). In the general case ($m\geq 1$) the greedy policy (for MD-MPI types~ 1 and~2) can be either directly estimated (Eq.~\eqref{eq:loss-trpo}) or trained to generalize the analytical solution (Eq.~\eqref{eq:loss-mappo}). The partial evaluation can be done using a TD-like approach, either done repeatedly while keeping the policy fixed or considering $m$-step rollouts.  Specifically, in the case of a KL divergence, one could use the fact that $\Omega_{\pi_k}^*(q_k(s,\cdot)) = \ln\sum_a \pi_k(a|s) \exp q_k(s,a)$ and that $\nabla \Omega_{\pi_k}^*(q_k(s,\cdot)) = \frac{\pi_k(\cdot|s)\exp q_k(s,\cdot)}{\sum_a \pi_k(a|s)\exp q_k(s,a)}$.

This general algorithmic scheme allows recovering state of the art algorithms. For example, MD-MPI type~2 with $m=\infty$ and a KL divergence as the regularizer is TRPO~\citep{schulman2015trust} (with a direct optimization of the regularized greedy step, as in Eq.~\eqref{eq:loss-trpo}, up to the use of a constraint instead of a regularization). DPP can be seen as a reparametrization\footnote{Indeed, if one see MD-MPI as a Mirror Descent approach, one can see DPP as a dual averaging approach, somehow updating a kind of cumulative q-functions directly in the dual. However, how to generalize this beyond the specific DPP algorithm is unclear, and we let it for future work.} of MD-MPI type~1 with $m=1$~\citep[Appx.~A]{azar2012dynamic}. MPO~\citep{abdolmaleki2018maximum} is derived from an expectation-maximization principle, but it can be seen as an instantiation of MD-MPI type~2, with a KL divergence, a greedy step similar to Eq.~\eqref{eq:loss-mappo} (up to additional regularization) and an evaluation step similar to Eq.~\eqref{eq:loss-sac} (without regularization, as in type~2, with m-step return and with the Retrace off-policy correction). This also generally applies to the approach proposed by~\citet{abdolmaleki2018relative} (up to an additional subtelty in the greedy step consisting in decoupling updates for the mean and variance in the case of a Gaussian policy).

\subsection{Analysis}

Here, we propose to analyze the error propagation of MD-MPI (and thus, its convergence and rate of convergence). We think this is an important topic, as it has only been partly studied for the special cases discussed in Sec.~\ref{subsec:related-algs-mdmpi}. For example, DPP enjoys an error propagation analysis in supremum norm (yet it is a reparametrization of a special case of MD-MPI, so not directly covered here), while TRPO or MPO are only guaranteed to have monotonic improvements, under some assumptions. Notice that we do not claim that our analysis covers all these cases, but it will provide the key technical aspects to analyze similar schemes (much like CBMPI compared to AMPI, as discussed in Sec.~\ref{subsec:analysis-regMPI} or by~\citet{scherrer2015approximate}; where the error is injected changes the bounds).

In Sec.~\ref{subsec:analysis-regMPI}, the analysis was a straightforward adaptation of the one of AMPI, thanks to the results of Sec.~\ref{sec:reg-mpds} (the regularized quantities behave like their unregularized counterparts). It is no longer the case here, as the regularizer changes over iterations, depending on what has been computed so far. We will notably need a slightly different notion of approximate regularized greediness.

\begin{definition}[Approximate Bregman divergence-regularized greediness]
    \label{def:approx-greedy-bregman}
    Write $J_k(\pi)$ the (negative) optimization problem corresponding to the Bregman divergence-regularized greediness (that is, negative regularized Bellman operator of $\pi$ applied to $v_k$):
    \begin{equation}
        J_k(\pi) = \langle-q_k, \pi\rangle + D_\Omega(\pi||\pi_k) = -T_{\pi,\Omega_{\pi_k}} v_k.
    \end{equation}
    We write $\pi_{k+1}\in\gc^{\epsilon'_{k+1}}_{\Omega_{\pi_k}}(v_k)$ if for any policy $\pi$ the policy $\pi_{k+1}$ satisfies
    \begin{equation}
        \langle \nabla J_k(\pi_{k+1}), \pi - \pi_{k+1}\rangle + \epsilon'_{k+1} \geq 0.
    \end{equation}
\end{definition}

In other words, $\pi_{k+1}\in\gc^{\epsilon'_{k+1}}_{\Omega_{\pi_k}}(v_k)$ means that $\pi_{k+1}$ is $\epsilon'_{k+1}$-close to satisfying the optimality condition, which might be slightly stronger than being $\epsilon'_{k+1}$-close to the optimal (as for AMPI or reg-MPI). Given this, we consider MD-MPI with errors in both greedy and evaluation steps, type~1
\begin{equation}
    \begin{cases}
        \pi_{k+1} = \gc^{\epsilon'_{k+1}}_{\Omega_{\pi_{k}}}(v_k)
        \\
        v_{k+1} = (T_{\pi_{k+1},\Omega_{\pi_{k}}})^m v_k + \epsilon_{k+1}
    \end{cases}
\end{equation}
and type~2
\begin{equation}
    \begin{cases}
        \pi_{k+1} = \gc^{\epsilon'_{k+1}}_{\Omega_{\pi_{k}}}(v_k)
        \\
        v_{k+1} = (T_{\pi_{k+1}})^m v_k + \epsilon_{k+1}
    \end{cases}.
\end{equation}

The quantity we are interested in is $v_* - v_{\pi_k}$, that is suboptimality in the unregularized MDP, while the algorithms compute new policies with a regularized greedy operator. So, we need to relate regularized and unregularized quantities when using a Bregman divergence based on the previous policy. The next lemma is the key technical result that allows analyzing MD-MPI.

\begin{lemma}
    \label{lemma:key-lemma}
    Assume that $\pi_{k+1}\in\gc^{\epsilon'_{k+1}}_{\Omega_{\pi_k}}(v_k)$, as defined in Def.~\ref{def:approx-greedy-bregman}. Then, the policy $\pi_{k+1}$ is $\epsilon'_{k+1}$-close to the regularized greedy policy, in the sense that for any policy $\pi$ %
    \begin{equation}
        T_{\pi,\Omega_{\pi_k}} v_k - T_{\pi_{k+1},\Omega_{\pi_k}} v_k \leq \epsilon'_{k+1}.
    \end{equation}
    Moreover, we can relate the (un)regularized Bellman operators applied to $v_k$. For any policy $\pi$ (so notably for the unregularized optimal policy $\pi_*$), we have
    \begin{align}
        T_{\pi} v_k - T_{\pi_{k+1},\Omega_{\pi_k}} v_k
        &\leq \epsilon'_{k+1} + D_\Omega(\pi||\pi_k) - D_\Omega(\pi||\pi_{k+1}),
        \\
        T_{\pi} v_k - T_{\pi_{k+1}} v_k
        &\leq \epsilon'_{k+1} + D_\Omega(\pi||\pi_k) - D_\Omega(\pi||\pi_{k+1}).
    \end{align}
\end{lemma}

We're interested in bounding the loss $l_k = v_* - v_{\pi_k}$, or some related quantity, for each type of MD-MPI. To do so, we introduce quantities similar to the ones of the AMPI analysis~\citep{scherrer2015approximate}, defined respectively for types~1 and~2: \textbf{1)} The distance between the optimal value function and the value before approximation at the $k^\text{th}$ iteration, $d_k^1 = v_* - (T_{\pi_k,\Omega_{\pi_{k-1}}})^m v_{k-1} = v_* - (v_k-\epsilon_k)$ and $d_k^2 = v_* - (T_{\pi_k})^m v_{k-1} = v_* - (v_k-\epsilon_k)$; \textbf{2)} The shift between the value before approximation and the policy value a iteration $k$, $s_k^1 = (T_{\pi_k,\Omega_{\pi_{k-1}}})^m v_{k-1} - v_{\pi_k} = (v_k-\epsilon_k) - v_{\pi_k}$ and $s_k^2 = (T_{\pi_k})^m v_{k-1} - v_{\pi_k} = (v_k-\epsilon_k) - v_{\pi_k}$; \textbf{3)} the Bellman residual at iteration $k$, $b^1_k = v_k - T_{\pi_{k+1},\Omega_{\pi_{k}}} v_k$ and $b^2_k = v_k - T_{\pi_{k+1}} v_k$.

For both types ($h\in\{1,2\}$), we have that $l_k^h = d_k^h + s_k^h$, so bounding the loss requires bounding these quantities, which is done in the following lemma (quantities related to both types enjoy the same bounds).

\begin{lemma}
    \label{lemma:bounds-bsd}
    Let $k\geq 1$, define $x_k = (I-\gamma P_{\pi_k}) \epsilon_k + \epsilon'_{k+1}$ and $y_k = -\gamma P_{\pi_*}\epsilon_k + \epsilon'_{k+1}$, as well as $\delta_k(\pi_*) = D_\Omega(\pi_*||\pi_k) - D_\Omega(\pi_*||\pi_{k+1})$. We have for $h\in\{1,2\}:$
    \begin{align}
        b^h_k &\leq (\gamma P_{\pi_k})^m b^h_{k-1} + x_k,
        \\
        s^h_k &\leq (\gamma P_{\pi_k})^m (I - \gamma P_{\pi_k})^{-1} b^h_{k-1} \text{ and}
        \\
        d^h_{k+1} &\leq \gamma P_{\pi_*} d_k^h + y_k + \sum_{j=1}^{m-1}(\gamma P_{\pi_{k+1}})^j b^h_k + \delta_k(\pi_*).%
    \end{align}
\end{lemma}

These bounds are almost the same as the ones of AMPI~\citep[Lemma~2]{scherrer2015approximate}, up to the additional $\delta_k(\pi_*)$ term in the bound of the distance $d_k^h$. One can notice that summing these terms gives a telescopic sum: $\sum_{k=0}^{K-1} \delta_k(\pi_*) = D_\Omega(\pi_*||\pi_0) - D_\Omega(\pi_*||\pi_{K})\leq D_\Omega(\pi_*||\pi_0)\leq \sup_\pi D_\Omega(\pi||\pi_0)$. For example, if $D_\Omega$ is the KL divergence and $\pi_0$ the uniform policy, then $\|\sup_\pi D_\Omega(\pi||\pi_0)\|_\infty = \ln|\A|$.   This suggests that
we must bound the regret $L_K$ defined as
\begin{equation}
    L_k = \sum_{k=1}^K l_k = \sum_{k=1}^K (v_* - v_{\pi_k}).
\end{equation}

\begin{theorem}
    \label{th:component-wise-md-mpi}
    Define $R_{\Omega_{\pi_0}}= \|\sup_{\pi} D_\Omega(\pi||\pi_0)\|_\infty$, after $K$ iterations of MD-MPI, for $h=1,2$, the regret satisfies
    \begin{align}
        L_K &\leq 2\sum_{k=2}^K\sum_{i=1}^{k-1} \sum_{j=i}^\infty \Gamma^j |\epsilon_{k-i}|
    + \sum_{k=1}^K \sum_{i=0}^{k-1} \sum_{j=i}^\infty \Gamma^j |\epsilon'_{k-i}|
    \\
    &\quad + \sum_{k=1}^K h(k)
    + \frac{1-\gamma^K}{(1-\gamma)^2} R_{\Omega_{\pi_0}} \un.
    \end{align}
    with $h(k) = 2\sum_{j=k}^\infty \Gamma^j |d_0|$ or $h(k) = 2\sum_{j=k}^\infty \Gamma^j |b_0|$.
\end{theorem}

From this,%
we can derive an $\ell_p$-bound for the regret.

\begin{cor}
    \label{cor:regret-md-mpi}
    Let $\rho$ and $\mu$ be distributions over states. Let $p$, $q$ and $q'$ be such that $\frac{1}{q} + \frac{1}{q'}=1$. Define the concentrability coefficients $C_q^i$ as in Cor.~\ref{cor:lp-bound-regMPI}. After $K$ iterations, the regret satisfies
    \begin{align}
        \|L_K\|_{p,\rho} &\leq  2\sum_{k=2}^K\sum_{i=1}^{k-1} \frac{\gamma^i}{1-\gamma} (C_q^i)^{\frac{1}{p}} \|\epsilon_{k-i}\|_{pq',\mu}
        \\ &\quad 
        + \sum_{k=1}^K\sum_{i=0}^{k-1} \frac{\gamma^i}{1-\gamma} (C_q^i)^{\frac{1}{p}} \|\epsilon'_{k-i}\|_{pq',\mu}
        \\ &\quad
        + g(k) + \frac{1-\gamma^K}{(1-\gamma)^2} R_{\Omega_{\pi_0}}.
    \end{align}
    with $g(k) = 2\sum_{k=1}^K \frac{\gamma^k}{1-\gamma} (C_q^k)^{\frac{1}{p}} \min(\|d_0\|_{pq',\mu}, \|b_0\|_{pq',\mu})$.
\end{cor}

This result bounds the regret, while it is usually the loss that is bounded. Both can be related as follows.

\begin{prop}
    \label{prop:loss-regret}
    For any $p\geq 1$ and distribution $\rho$, we have %
    $%
        \min_{1\leq k \leq K} \|v_* - v_{\pi_k}\|_{1,\rho} \leq \frac 1 K \|L_K\|_{p,\rho}.
    $%
\end{prop}

This means that if we can control the average regret, then we can control the loss of the best policy computed so far. This suggests that practically we should not use the last policy, but this best policy.%

From Cor.~\ref{cor:regret-md-mpi} can be derived the convergence and rate of convergence of MD-MPI in the exact case.

\begin{cor}
    \label{cor:regret-no-error}
    Both MD-MPI type~1 and~2 enjoy the following rate of convergence, when no approximation is done ($\epsilon_k = \epsilon'_k = 0$),
    \begin{equation}
        \frac 1 K \|L_K\|_\infty \leq \frac{1-\gamma^K}{(1-\gamma)^2} \frac{2 \gamma \|v_* - v_0\|_\infty + R_{\Omega_{\pi_0}}}{K}.
    \end{equation}
\end{cor}

In classical DP and in regularized DP (see Cor.~\ref{cor:lp-bound-regMPI}), there is a linear convergence rate (the bound is $\frac{2\gamma^K}{1-\gamma}\|v_*-v_0\|_\infty$), while in this case we only have a logarithmic convergence rate. We also pay an horizon factor (square dependency in $\frac{1}{1-\gamma}$ instead of linear). This is normal, as we bound the regret instead of the loss. Bounding the regret in classical DP would lead to the bound of Cor.~\ref{cor:regret-no-error} (without the $R_{\Omega_{\pi_0}}$ term).

The convergence rate of the loss of MD-MPI is an open question, but a sublinear rate is quite possible.
Compared to classical DP, we slow down greediness by adding the Bregman divergence penalty. 
Yet, this kind of regularization is used in an approximate setting, where it favors stability empirically (even if studying this further would require much more work regarding the $\|\epsilon'_k\|$ term, as discussed in Sec.~\ref{subsec:analysis-regMPI}).

As far as we know, the only other approach that studies a DP scheme regularized by a divergence and that offers a convergence rate is DPP, up to the reparameterization we discussed earlier. MD-MPI has the same upper-bound as DPP in the exact case~\citep[Thm.~2]{azar2012dynamic}. However, DPP bounds the loss, while we bound a regret. This means that if the rate of convergence of our loss can be sublinear, it is superlogarithmic (as the rate of the regret is logarithmic), while the rate of the loss of DPP is logarithmic.

To get more insight on Cor.~\ref{cor:regret-md-mpi}, we can group the terms differently, by grouping the errors.

\begin{cor}
    \label{cor:sum-errors}
    With the same notations as Cor.~\ref{cor:regret-md-mpi}, we have
    \begin{align}
        \frac{1}{K} \|L_k\|_{p,\rho} &\leq \sum_{i=1}^{K-1} \frac{\gamma^i}{1-\gamma} (C_q^i)^{\frac 1 p} \frac{2 E_{K-i} + E'_{K-i}}{K}
        \\
        &\quad + \frac{1}{K} \left(g(k) + \frac{1-\gamma^K}{(1-\gamma)^2} R_{\Omega_{\pi_0}}\right),
    \end{align}
    with $E_i = \sum_{j=1}^i \|\epsilon_j\|_{pq',\mu}$ and $E'_i = \sum_{j=1}^i \|\epsilon'_j\|_{pq',\mu}$.
\end{cor}

Compared to the bound of AMPI~\citep[Thm.~7]{scherrer2015approximate}, instead of propagating the errors, we propagate the sum of errors over previous iterations normalized by the total number of iterations. So, contrary to approximate DP, it is no longer the last iterations that have the highest influence on the regret. Yet, we highlight again the fact that we bound a regret, and bounding the regret of AMPI would provide a similar result.

Our result is similar to the error propagation of DPP~\citep[Thm.~5]{azar2012dynamic}, except that we sum norms of errors, instead of norming a sum of errors, the later being much better (as it allows the noise to cancel over iterations). Yet, as said before, DPP is not a special case of our framework, but a reparameterization of such one. Consequently, while we estimate value functions, DPP estimate roughly at iteration $k$ a sum of $k$ advantage functions (converging to $-\infty$ for any suboptimal action in the exact case). As explained before, where the error is injected does matter. Knowing if the DPP's analysis can be generalized to our framework (MPI scheme, $\ell_p$ bounds) remains an open question.

To get further insight, we can express the bound using different concentrability coefficients.

\begin{cor}
    \label{cor:group-c}
    Define the concentrability coefficient $C_q^{l,k}$ as
    $
        C_q^{l,k} = \frac{(1-\gamma)^2}{\gamma^l - \gamma^k} \sum_{i=l}^{k-1}\sum_{j=i}^\infty c_q(j)
    $,
    the regret then satisfies
    \begin{align}
        \|L_K\|_{p,\rho} &\leq 2 \sum_{i=1}^{K-1} \frac{\gamma -\gamma^{i+1}}{(1-\gamma)^2} (C_q^{1,i+1})^{\frac 1 p} \|\epsilon_{K-i}\|_{pq',\mu}
        \\ 
        &\quad +
        \sum_{i=0}^{K-1} \frac{1 -\gamma^{i+1}}{(1-\gamma)^2} (C_q^{0,i+1})^{\frac 1 p } \|\epsilon'_{K-i}\|_{pq',\mu} + f(k)
    \end{align}
    with $f(k) = \frac{\gamma -\gamma^{K+1}}{(1-\gamma)^2} (C_q^{1,K+1})^{\frac 1 p}\min(\|d_0\|_{pq',\mu},\|b_0\|_{pq',\mu}) + \frac{1-\gamma^K}{(1-\gamma)^2} R_{\Omega_{\pi_0}}$.
\end{cor}

We observe again that contrary to ADP, the last iteration does not have the highest influence, and we do not enjoy a decrease of influence at the exponential rate $\gamma$ towards the initial iterations. However, we bound a different quantity (regret instead of loss), that explains  this behavior. Here again, bounding the regret in AMPI would lead to the same bound (up to the term $R_{\Omega_{\pi_0}}$). Moreover, sending $p$ and $K$ to infinity, defining $\epsilon = \sup_j \|\epsilon_j\|_\infty$ and $\epsilon' = \sup_j \|\epsilon'_j\|_\infty$, we get
$
    \limsup\limits_{K\rightarrow\infty} \frac 1 K \|L_K\|_\infty \leq \frac{2\gamma \epsilon +  \epsilon'}{(1-\gamma)^2}
$, which is the classical asymptotical bound for approximate value and policy iterations~\citep{Bertsekas:1996:NP:560669} (usually stated without greedy error). It is generalized here to an approximate MPI scheme regularized with a Bregman divergence.

\section{Conclusion}
\label{sec:conclusion}

We have introduced a general theory of regularized MDPs, where the usual Bellman evaluation operator is modified by either a fixed convex function or a Bregman divergence between consecutive policies. For both cases, we proposed a general algorithmic scheme based on MPI. We shown how many (variations of) existing algorithms could be derived from this general algorithmic scheme, and also analyzed and discussed the related propagation of errors. 

We think that this framework can open many perspectives, among which links between (approximate) DP and proximal convex optimization (going beyond mirror descent), temporal consistency equations (roughly regularized Bellman residuals), regularized policy search (maximizing the expected regularized value function), inverse reinforcement learning (thanks to uniqueness of greediness in this regularized framework) or zero-sum Markov games (regularizing the two-player Bellman operators). We develop more these points in the appendix.
 
This work also lefts open questions, such as combining the propagation of errors with a finite sample analysis, or what specific regularizer one should choose for what context.
Some approaches also combine a fixed regularizer and a divergence~\citep{akrour2018model}, a case not covered here and worth being investigated.

\clearpage 

\bibliography{bib}
\bibliographystyle{icml2019}

\clearpage
\onecolumn

\appendix

This appendices provide the proofs for all stated results (Appx.~\ref{appx:proofs-A} to~\ref{appx:proofs-C}) and discuss in more details the perspectives mentioned in Sec.~\ref{sec:mdmpi} (Appx.~\ref{appx:discussion-perspectives}).

\section{Proofs of section~\ref{sec:reg-mpds}}
\label{appx:proofs-A}

In this section, we prove the results of Sec.~\ref{sec:reg-mpds}. We start with the properties of the regularized Bellman operators.

\begin{proof}[Proof of Proposition~\ref{prop:reg-op}]
    We can write $T_{\pi,\Omega}v = r_\pi - \Omega(\pi) + \gamma P_\pi v$, it is obviously affine (in $v$). 
    Then, we show that the operators are monotonous. For the evaluation operator, we have
    \begin{equation}
        v_1\geq v_2 \Rightarrow T_{\pi} v_1 \geq T_{\pi} v_2 \Leftrightarrow T_{\pi,\Omega} v_1 = T_\pi v_1 - \Omega(\pi) \geq T_\pi v_2 - \Omega(\pi) = T_{\pi,\Omega} v_2.
    \end{equation}
    For the optimality operator, we have
    \begin{align}
        v_1 \geq v_2 &\Rightarrow \forall s\in\s, \quad q_{s,1} \geq q_{s,2} & &
        \\
        &\Rightarrow \forall s\in\s, \quad \Omega^*(q_{s,1}) \geq \Omega^*(q_{s,2}) & & \text{by Prop.~\ref{prop:convex-conjugate}}
        \\
        &\Leftrightarrow \forall s\in\s, \quad  [T_{*,\Omega}v_1](s) \geq  [T_{*,\Omega}v_2](s) & & 
        \\
        &\Leftrightarrow  T_{*,\Omega}v_1 \geq  T_{*,\Omega}v_2. & &
    \end{align}
    Then, we show the distributivity property. For the evaluation operator, we have
    \begin{equation}
        T_{\pi,\Omega}(v+c\un) = T_\pi (v+c\un) - \Omega(\pi) = T_\pi v + \gamma c \un - \Omega(\pi) = T_{\pi,\Omega} v + \gamma c \un. 
    \end{equation}
    For the optimality operator, for any $s\in\s$, we have
    \begin{align}
         [T_{*,\Omega}(v+c\un)](s) &= \Omega^*(q_s + \gamma c \un) & & 
         \\
         &= \Omega^*(q_s) + \gamma c & &\text{by Prop.~\ref{prop:convex-conjugate}}
         \\
         &= [T_{*,\Omega}v](s) + \gamma c. & & 
    \end{align}
    Lastly, we study the contraction of both operators. For the evaluation operator, we have
    \begin{equation}
        T_{\pi,\Omega}v_1 - T_{\pi,\Omega}v_2 = T_{\pi}v_1 - \Omega(\pi) - \left(T_{\pi}v_2 - \Omega(\pi)\right) = T_{\pi}v_1 - T_{\pi}v_2.
    \end{equation}
    So, the contraction is the same as the one of the unregularized operator.
    For the optimality operator, we have that 
    \begin{equation}
         \left\|T_{*,\Omega}v_1 - T_{*,\Omega}v_2\right\|_\infty = \max_{s\in\s} \left|[T_{*,\Omega}v_1](s) - [T_{*,\Omega}v_2](s)\right|.
    \end{equation}
    Pick $s\in\s$, and without loss of generality assume that $[T_{*,\Omega}v_1](s) \geq [T_{*,\Omega}v_2](s)$. Write also $\pi_1=\g_\Omega(v_1)$ and $\pi_2=\g_\Omega(v_2)$. We have
    \begin{align}
        |[T_{*,\Omega}v_1](s) - [T_{*,\Omega}v_2](s)| &= [T_{*,\Omega}v_1](s) - [T_{*,\Omega}v_2](s) & &
        \\
        &= [T_{\pi_1,\Omega}v_1](s) - [T_{\pi_2,\Omega}v_2](s) & &
        \\
        &\leq [T_{\pi_1,\Omega}v_1](s) - [T_{\pi_1,\Omega}v_2](s) & &\text{as } T_{*,\Omega} v_2 = T_{\pi_2,\Omega} v_2 \geq T_{\pi_1,\Omega} v_2,
        \\
        &\leq \gamma \|v_1- v_2 \|_\infty. & &
    \end{align}
    The stated result follows immediately.
\end{proof}

Then, we show that in a regularized MDP, the policy greedy respectively to the optimal value function is indeed the optimal policy, and is unique.

\begin{proof}[Proof of Theorem~\ref{th:opt-reg-pol}]
    The uniqueness of $\pi_{*,\Omega}$ is a consequence of the strong convexity of $\Omega$, see Prop.~\ref{prop:convex-conjugate}.
    On the other hand, by definition of the greediness, we have
    \begin{equation}
        \pi_{*,\Omega} = \g_{\Omega}(v_{*,\Omega}) \Leftrightarrow T_{\pi_{*,\Omega},\Omega} v_{*,\Omega} = T_{*,\Omega} v_{*,\Omega} = v_{*,\Omega}.
    \end{equation}
    This proves that $v_{*,\Omega}$ is the value function of $\pi_{*,\Omega}$. Next, for any function $v\in\R^\s$ and any policy $\pi$, we have
    \begin{equation}
        T_{*,\Omega} v \geq T_{\pi,\Omega} v.
    \end{equation}
    Using monotonicity, we have that
    \begin{equation}
        T_{*,\Omega}^2 v = T_{*,\Omega}(T_{*,\Omega} v)
        \geq T_{*,\Omega}(T_{\pi,\Omega} v)
        \geq T_{\pi,\Omega}(T_{\pi,\Omega} v) = T^2_{\pi,\Omega} v.
    \end{equation}
    By direct induction, for any $n\geq 1$, $T_{*,\Omega}^n v \geq T^n_{\pi,\Omega} v$. Taking the limit as $n\rightarrow \infty$, we conclude that $v_{*,\Omega}\geq v_{\pi,\Omega}$.
\end{proof}

Next, we relate regularized and unregularized value functions (for a given policy, and for the optimal value function).

\begin{proof}[Proof of Proposition~\ref{prop:bound-values-reg-unreg}]
    We start by linking (un)regularized values of a given policy.
    Let $v\in\R^\s$. As $T_{\pi,\Omega}v = T_\pi v - \Omega(\pi)$, we have that
    \begin{equation}
        T_{\pi} v - U_\Omega\un \leq T_{\pi,\Omega} v \leq T_{\pi} v - L_\Omega \un.  
    \end{equation}
    We work on the left inequality first. We have
    \begin{equation}
        T_{\pi,\Omega}^2 v = T_{\pi,\Omega}(T_{\pi,\Omega} v)
        \geq T_{\pi,\Omega}(T_\pi v - U_\Omega\un)
        \geq T_{\pi}(T_\pi v - U_\Omega\un) - U_\Omega \un 
        = T_\pi^2 v - \gamma U_\Omega \un - U_\Omega\un.
    \end{equation}
    By direct induction, for any $n\geq 1$,
    \begin{equation}
        T_{\pi,\Omega}^n \geq  T_\pi^n v - \sum_{k=0}^{n-1}\gamma^k U_\Omega \un.
    \end{equation}
    Taking the limit as $n\rightarrow \infty$ we obtain
    \begin{equation}
        v_{\pi,\Omega} \geq v_{\pi} -\frac{U_\Omega}{1-\gamma}\un.
    \end{equation}
    The proof is similar for the right inequality.
    Next, we link the (un)regularized optimal values.
    As a direct corollary of Prop.~\ref{prop:convex-conjugate}, for any $v\in\R^\s$ we have
    \begin{equation}
        T_* v - U_\Omega\un \leq T_{*,\Omega} v \leq T_* v - L_\Omega\un.
    \end{equation}
    Then, the proof is the same as above, switching evaluation and optimality operators.
\end{proof}

Lastly, we show how good is the optimal policy of the regularized MDP for the original problem (unregularized MDP).

\begin{proof}[Proof of Theorem~\ref{th:reg-opt-in-original-mdp}]
    The right inequality is obvious, as for any $\pi$, $v_* \geq v_\pi$. For the left inequality,
    \begin{align}
        v_* &\leq v_{*,\Omega} + \frac{U_\Omega}{1-\gamma} & &\text{by Prop.~\ref{prop:bound-values-reg-unreg}}
        \\
        &= v_{\pi_{*,\Omega},\Omega} + \frac{U_\Omega}{1-\gamma} & &\text{by Thm.~\ref{th:opt-reg-pol}}
        \\
        &\leq v_{\pi_{*,\Omega}} + \frac{U_\Omega}{1-\gamma} - \frac{L_\Omega}{1-\gamma} & &\text{by Prop.~\ref{prop:bound-values-reg-unreg}}.
    \end{align}
\end{proof}

\section{Proofs of section~\ref{sec:regmpi}}
\label{appx:proofs-B}

The results of section~\ref{sec:regmpi} do not need to be proven. Indeed, we have shown in Sec.~\ref{sec:reg-mpds} that all involved quantities of regularized MDPs satisfy the same properties as their unregularized counterpart. Therefore, the proofs of these results are identical to the proofs provided by~\citet{scherrer2015approximate}, up to the replacement of value functions, Bellman operators, and so on, by their regularized counterparts. The proofs for Mirror Descent Modified Policy Iteration (Sec.~\ref{sec:mdmpi}) are less straightforward.

\section{Proofs of section~\ref{sec:mdmpi}}
\label{appx:proofs-C}

As a prerequisite of Lemma~\ref{lemma:key-lemma}, we need the following result.

\begin{lemma}[Three-point identity]
    \label{lemma:3point}
    Let $\pi$ be any policy, we have that
    \begin{equation}
        \langle \nabla\Omega(\pi_k) - \nabla\Omega(\pi_{k+1}),\pi - \pi_{k+1}\rangle
        = D_\Omega(\pi||\pi_{k+1}) - D_\Omega(\pi||\pi_{k}) + D_\Omega(\pi_{k+1}||\pi_k).
    \end{equation}
\end{lemma}
\begin{proof}
    This is the classical three-point identity of Bregman divergences, and can be checked by calculus:
    \begin{align}
        D_\Omega(\pi||\pi_{k+1}) - D_\Omega(\pi||\pi_{k}) + D_\Omega(\pi_{k+1}||\pi_k)
        &= \Omega(\pi) - \Omega(\pi_{k+1}) - \langle\nabla\Omega(\pi_{k+1}), \pi - \pi_{k+1}\rangle
        \\
        &\quad - \Omega(\pi) + \Omega(\pi_{k}) + \langle\nabla\Omega(\pi_{k}), \pi - \pi_{k}\rangle
        \\
        &\quad + \Omega(\pi_{k+1}) - \Omega(\pi_{k}) - \langle\nabla\Omega(\pi_{k}), \pi_{k+1} - \pi_{k}\rangle
        \\
        &= \langle \nabla\Omega(\pi_k) - \nabla\Omega(\pi_{k+1}),\pi - \pi_{k+1}\rangle.
    \end{align}
\end{proof}

Now, we can prove the key lemma of MD-MPI.

\begin{proof}[Proof of Lemma~\ref{lemma:key-lemma}]
    Let $J_k$ be as defined in Def.~\ref{def:approx-greedy-bregman},
    \begin{equation}
        J_k(\pi) = \langle-q_k, \pi\rangle + D_\Omega(\pi||\pi_k) = -T_{\pi,\Omega_{\pi_k}} v_k,
    \end{equation}
    and let $\pi_{k+1}\in\gc^{\epsilon'_{k+1}}_{\Omega_{\pi_k}}(v_k)$, that is $\langle \nabla J_k(\pi_{k+1}), \pi - \pi_{k+1}\rangle + \epsilon'_{k+1} \geq 0$. By convexity of $J_k$, for any policy $\pi$, we have
    \begin{align}
        J_k(\pi) - J_k(\pi_{k+1}) &\geq \langle\nabla J(\pi_{k+1}), \pi - \pi_{k+1}\rangle & &\text{by convexity of $J_k$} 
        \\
        &\geq - \epsilon'_{k+1} & &\text{as $\pi_{k+1}\in\gc^{\epsilon'_{k+1}}_{\Omega_{\pi_k}}(v_k)$}
        \\ \Leftrightarrow
        - T_{\pi,\Omega_{\pi_k}} v_k + T_{\pi_{k+1},\Omega_{\pi_k}} v_k &\geq - \epsilon'_{k+1} & &\text{using $J_k(\pi)=-T_{\pi,\Omega_{\pi_k}} v_k$}
        \\ \Leftrightarrow
        T_{\pi,\Omega_{\pi_k}} v_k - T_{\pi_{k+1},\Omega_{\pi_k}} v_k  &\leq \epsilon'_{k+1} & &
    \end{align}
    This is the first result stated in Lemma~\ref{lemma:key-lemma}.
    
    Next, we relate (un)regularized quantities. We start with the following decomposition
    \begin{equation}
        \langle -q_k, \pi\rangle = \langle - q_k, \pi_{k+1} \rangle + \langle -q_k, \pi - \pi_{k+1}\rangle.
        \label{proof:lemma-key:decomposition}
    \end{equation}
    Taking the gradient of $J_k$ (by using the definition of the Bregman divergence), we get
    \begin{align}
        \nabla J_k(\pi_{k+1}) &= -q_k + \nabla\Omega(\pi_{k+1}) - \nabla \Omega(\pi_k)
        \\
        \Leftrightarrow
        -q_k &= \nabla J_k(\pi_{k+1}) + \nabla\Omega(\pi_k) - \nabla\Omega(\pi_{k+1}).
        \label{proof:lemma-key:gradient}
    \end{align}
    Injecting Eq.~\eqref{proof:lemma-key:gradient} into Eq.~\eqref{proof:lemma-key:decomposition}, we get
    \begin{align}
        \langle -q_k, \pi\rangle &= \langle - q_k, \pi_{k+1} \rangle + \langle \nabla J_k(\pi_{k+1}) + \nabla\Omega(\pi_k) - \nabla\Omega(\pi_{k+1}), \pi - \pi_{k+1}\rangle
        \\
        &= \langle - q_k, \pi_{k+1} \rangle + \underbrace{\langle \nabla J_k(\pi_{k+1}),\pi - \pi_{k+1}\rangle}_{\geq - \epsilon'_{k+1}} + \underbrace{\langle \nabla\Omega(\pi_k) - \nabla\Omega(\pi_{k+1}), \pi - \pi_{k+1}\rangle}_{\text{three-point identity (Lemma~\ref{lemma:3point})}}
        \\
         &\geq \langle - q_k, \pi_{k+1} \rangle - \epsilon'_{k+1} +  D_\Omega(\pi||\pi_{k+1}) - D_\Omega(\pi||\pi_{k}) + D_\Omega(\pi_{k+1}||\pi_k)
        \\
        \Leftrightarrow
        T_{\pi} v_k &\leq T_{\pi_{k+1}} v_k + \epsilon'_{k+1} + D_\Omega(\pi||\pi_{k}) -  D_\Omega(\pi||\pi_{k+1}) - D_\Omega(\pi_{k+1}||\pi_k), \label{proof:lemma-key:almost-end}
    \end{align}
    where we used in the last inequality the fact that $\langle q_k, \pi\rangle= T_\pi v_k$. From the definition of the regularized Bellman operator, we have that $T_{\pi_{k+1}} v_k - D_\Omega(\pi_{k+1}||\pi_k) = T_{\pi_{k+1}, \Omega_{\pi_k}} v_k$, so Eq.~\eqref{proof:lemma-key:almost-end} is equivalent to the second result of Lemma~\ref{lemma:key-lemma}:
    \begin{equation}
        T_{\pi} v_k \leq T_{\pi_{k+1}, \Omega_{\pi_k}} v_k + \epsilon'_{k+1} + D_\Omega(\pi||\pi_{k}) -  D_\Omega(\pi||\pi_{k+1}).
    \end{equation}
    As the Bregman divergence is positive, $-D_\Omega(\pi||\pi_{k+1})\leq 0$, and thus Eq.~\eqref{proof:lemma-key:almost-end} implies the last result of Lemma~\ref{lemma:key-lemma}:
    \begin{equation}
        T_{\pi} v_k \leq T_{\pi_{k+1}} v_k + \epsilon'_{k+1} + D_\Omega(\pi||\pi_{k}) -  D_\Omega(\pi||\pi_{k+1}).
    \end{equation}
    This concludes the proof.
\end{proof}

Next, we prove the bounds for $b^h_k$, $s^h_k$ and $d^h_k$, for $h=1,2$ (if the bounds are the same, the proofs differ).

\begin{proof}[Proof of Lemma~\ref{lemma:bounds-bsd}]
    We start by bounding the quantities for MD-MPI type~1. First, we consider the Bellman residual:
    \begin{align}
        b^1_k &= v_k - T_{\pi_{k+1},\Omega_{\pi_k}} v_k
        \\
        &\myeq{a}{=} v_k - T_{\pi_k} v_k + T_{\pi_{k},\Omega_{\pi_k}} v_k - T_{\pi_{k+1},\Omega_{\pi_k}} v_k
        \\
        &\myeq{b}{\leq} v_k - T_{\pi_k,\Omega_{\pi_{k-1}}} v_k + \epsilon'_{k+1}
        \\
        &=  v_k - \epsilon_k - T_{\pi_k,\Omega_{\pi_{k-1}}} v_k + \gamma P_{\pi_k}\epsilon_k + \epsilon_k - \gamma P_{\pi_k}\epsilon_k + \epsilon'_{k+1}
        \\
        &\myeq{c}{=} (v_k - \epsilon_k) - T_{\pi_k,\Omega_{\pi_{k-1}}}(v_k - \epsilon_k) + x_k
        \\
        &\myeq{d}{=} (T_{\pi_k,\Omega_{\pi_{k-1}}})^m v_{k-1} - T_{\pi_k,\Omega_{\pi_{k-1}}}(T_{\pi_k,\Omega_{\pi_{k-1}}})^{m} v_{k-1} + x_k
        \\
        &= (T_{\pi_k,\Omega_{\pi_{k-1}}})^m v_{k-1} - (T_{\pi_k,\Omega_{\pi_{k-1}}})^m T_{\pi_k,\Omega_{\pi_{k-1}}} v_{k-1} + x_k
        \\
        &= (\gamma P_{\pi_{k}})^m (v_{k-1} - T_{\pi_k,\Omega_{\pi_{k-1}}} v_{k-1}) + x_k
        \\
        &= (\gamma P_{\pi_{k}})^m b^1_{k-1} + x_k.
    \end{align}
    In the previous equations, we used the following facts:
    \begin{enumerate}[(a)]
        \item $T_{\pi_k} v_k = T_{\pi_{k},\Omega_{\pi_k}} v_k$ as $\Omega_{\pi_k}(\pi_k)=0$.
        \item We used two facts. First, $T_{\pi_k} v_k \geq T_{\pi_k} v_k - \Omega_{\pi_{k-1}}(\pi_k) = T_{\pi_k,\Omega_{\pi_{k-1}}} v_k$, as $\Omega_{\pi_{k-1}}(\pi_k)\geq 0$. Second, by Lemma~\ref{lemma:key-lemma}, $T_{\pi_{k},\Omega_{\pi_k}} v_k - T_{\pi_{k+1},\Omega_{\pi_k}} v_k \leq \epsilon'_{k+1}$.
        \item We used two facts. First, generally speaking, we have $T_{\pi,\Omega}(v_1 + v_2) = T_{\pi,\Omega} v_1 + \gamma P_\pi v_2$ (as $T_{\pi,\Omega}$ is affine). Second, by definition $x_k=(I- \gamma P_{\pi_k})\epsilon_k + \epsilon'_{k+1}$.
        \item By definition, $v_k = (T_{\pi_k,\Omega_{\pi_{k-1}}})^m v_{k-1} + \epsilon_k$.
    \end{enumerate}
    
    Next, we bound the shift $s_k^1$:
    \begin{align}
        s_k^1 &= (T_{\pi_k,\Omega_{\pi_{k-1}}})^m v_{k-1} - v_{\pi_k}
        \\
        &= (T_{\pi_k,\Omega_{\pi_{k-1}}})^m v_{k-1} - v_{\pi_k, \Omega_{\pi_{k-1}}} + v_{\pi_k, \Omega_{\pi_{k-1}}} - v_{\pi_k}
        \\
        &\myeq{a}{\leq} (T_{\pi_k,\Omega_{\pi_{k-1}}})^m v_{k-1} - v_{\pi_k, \Omega_{\pi_{k-1}}}
        \\
        &\myeq{b}{=} (T_{\pi_k,\Omega_{\pi_{k-1}}})^m v_{k-1} - (T_{\pi_k,\Omega_{\pi_{k-1}}})^\infty v_{k-1}
        \\
        &= (T_{\pi_k,\Omega_{\pi_{k-1}}})^m v_{k-1} - (T_{\pi_k,\Omega_{\pi_{k-1}}})^m (T_{\pi_k,\Omega_{\pi_{k-1}}})^\infty v_{k-1}
        \\
        &= (\gamma P_{\pi_k})^m (v_{k-1} - (T_{\pi_k,\Omega_{\pi_{k-1}}})^\infty v_{k-1})
        \\
        &= (\gamma P_{\pi_k})^m \sum_{j=0}^\infty ((T_{\pi_k,\Omega_{\pi_{k-1}}})^j v_{k-1} - (T_{\pi_k,\Omega_{\pi_{k-1}}})^{j+1} v_{k-1})
        \\
        &= (\gamma P_{\pi_k})^m \sum_{j=0}^\infty ((T_{\pi_k,\Omega_{\pi_{k-1}}})^j v_{k-1} - (T_{\pi_k,\Omega_{\pi_{k-1}}})^{j} T_{\pi_k,\Omega_{\pi_{k-1}}} v_{k-1})
        \\
        &= (\gamma P_{\pi_k})^m \sum_{j=0}^\infty (\gamma P_{\pi_k})^j (v_{k-1} - T_{\pi_k,\Omega_{\pi_{k-1}}} v_{k-1})
        \\
        &= (\gamma P_{\pi_k})^m (I - \gamma P_{\pi_k})^{-1} b^1_{k-1}.
    \end{align}
    In the previous equations, we used the following facts:
    \begin{enumerate}[(a)]
        \item Generally speaking, if $\Omega\geq 0$ then $v_{\pi,\Omega} - v_\pi\leq 0$ (see Prop.~\ref{prop:bound-values-reg-unreg}).
        \item With a slight abuse of notation, for any $v\in\R^\s$, $v_{\pi,\Omega} = T_{\pi,\Omega}^\infty v$.
    \end{enumerate}
    
    Then we bound the distance $d_k^1$:
    \begin{align}
        d_{k+1}^1 &= v_* - (T_{\pi_{k+1},\Omega_{\pi_k}})^m v_k
        \\
        &= T_{\pi_*} v_* - T_{\pi_*} v_k + T_{\pi_*} v_k - T_{\pi_{k+1},\Omega_{\pi_k}} v_k + T_{\pi_{k+1},\Omega_{\pi_k}} v_k - (T_{\pi_{k+1},\Omega_{\pi_k}})^m v_k
        \\
        &= \gamma P_{\pi_*} (v_* - v_k) + T_{\pi_*} v_k - T_{\pi_{k+1},\Omega_{\pi_k}} v_k + T_{\pi_{k+1},\Omega_{\pi_k}} v_k - (T_{\pi_{k+1},\Omega_{\pi_k}})^m v_k
        \\
        &\myeq{a}{\leq} \gamma P_{\pi_*} (v_* - v_k) + \epsilon'_{k+1} + \delta_k(\pi_*) + T_{\pi_{k+1},\Omega_{\pi_k}} v_k - (T_{\pi_{k+1},\Omega_{\pi_k}})^m v_k 
        \\
        &\myeq{b}{=}  \gamma P_{\pi_*} (v_* - v_k) + \epsilon'_{k+1} + \delta_k(\pi_*) + \sum_{j=1}^{m-1} (\gamma P_{\pi_{k+1}})^j b_k^1
        \\
        &= \gamma P_{\pi_*} (v_* - v_k) + \gamma P_{\pi_*} \epsilon_k - \gamma P_{\pi_*} \epsilon_k  + \epsilon'_{k+1} + \delta_k(\pi_*) + \sum_{j=1}^{m-1} (\gamma P_{\pi_{k+1}})^j b_k^1
        \\
        &\myeq{c}{=} \gamma P_{\pi_*} d^1_k + y_k + \delta_k(\pi_*) + \sum_{j=1}^{m-1} (\gamma P_{\pi_{k+1}})^j b_k^1.
    \end{align}
    In the previous equations, we used the following facts:
    \begin{enumerate}[(a)]
        \item By Lemma~\ref{lemma:key-lemma},
        \begin{equation}
            T_{\pi_*} v_k - T_{\pi_{k+1},\Omega_{\pi_k}} v_k \leq \epsilon'_{k+1} + \delta_k(\pi_*).
        \end{equation}
        \item For this step, we used:
        \begin{align}
            T_{\pi_{k+1},\Omega_{\pi_k}} v_k - (T_{\pi_{k+1},\Omega_{\pi_k}})^m v_k
            &= \sum_{j=1}^{m-1} ((T_{\pi_{k+1},\Omega_{\pi_k}})^j v_k - (T_{\pi_{k+1},\Omega_{\pi_k}})^{j+1} v_k)
            \\
            &= \sum_{j=1}^{m-1} ((T_{\pi_{k+1},\Omega_{\pi_k}})^j v_k - (T_{\pi_{k+1},\Omega_{\pi_k}})^{j} T_{\pi_{k+1},\Omega_{\pi_k}} v_k)
            \\
            &= \sum_{j=1}^{m-1} (\gamma P_{\pi_{k+1}})^j (v_k - T_{\pi_{k+1},\Omega_{\pi_k}} v_k)
            \\
            &= \sum_{j=1}^{m-1} (\gamma P_{\pi_{k+1}})^j b_k^1
        \end{align}
        \item By definition of $d^1_k = v_* - (v_k - \epsilon_k)$ and $y_k = - \gamma P_{\pi_*} \epsilon_k  + \epsilon'_{k+1}$.
    \end{enumerate}
    
    The proofs for the quantities involved in MD-MPI type~2 are similar, even if their definition differ. First, we consider the Bellman residual
    \begin{align}
        b_k^2 &= v_k - T_{\pi_{k+1}} v_k
        \\
        &\myeq{a}{\leq} v_k - T_{\pi_{k+1}} v_k + \Omega_{\pi_k}(\pi_{k+1}) = v_k - T_{\pi_{k+1},\Omega_{\pi_k}} v_k
        \\
        &\myeq{b}{=} v_k - T_{\pi_{k}} v_k + T_{\pi_{k},\Omega_{\pi_k}} v_k - T_{\pi_{k+1},\Omega_{\pi_k}} v_k
        \\
        &\myeq{c}{\leq} v_k - T_{\pi_{k}} v_k + \epsilon'_{k+1}
        \\
        &= v_k-\epsilon_k -T_{\pi_k} v_k + \gamma P_{\pi_k} \epsilon_k + \epsilon_k - \gamma P_{\pi_k} \epsilon_k + \epsilon'_{k+1}
        \\
        &\myeq{d}{=} (v_k-\epsilon_k) -T_{\pi_k} v_k (v_k-\epsilon_k) + x_k
        \\
        &\myeq{e}{=} (T_{\pi_{k}})^m v_{k-1} - T_{\pi_{k}} (T_{\pi_{k}})^m v_{k-1} + x_k = (\gamma P_{\pi_k})^m (v_{k-1} - T_{\pi_{k}}v_{k-1}) + x_k = (\gamma P_{\pi_k})^m b_{k-1}^2 + x_k.
    \end{align}
    In the previous equations, we used the following facts:
    \begin{enumerate}[(a)]
        \item This is because $\Omega_{\pi_k}(\pi_{k+1})\geq 0$ and by definition of $T_{\pi,\Omega} v = T_\pi v - \Omega(\pi)$.
        \item It uses the fact that $T_{\pi_{k}} v_k  = T_{\pi_{k},\Omega_{\pi_k}} v_k$ (as $\Omega_{\pi_k}(\pi_k)=0$).
        \item This is by Lemma~\ref{lemma:key-lemma}.
        \item This is by definition of $x_k = \epsilon_k - \gamma P_{\pi_k} \epsilon_k + \epsilon'_{k+1}$.
        \item This is by definition of $v_k = (T_{\pi_{k}})^m v_{k-1} + \epsilon_k$.
    \end{enumerate}
    
    Then, we bound the shift $s_k^2$, the technique being the same as before:
    \begin{align}
        s_k^2 &= (T_{\pi_k})^m v_{k-1} - v_{\pi_k}
        \\
        &= (T_{\pi_k})^m v_{k-1} - (T_{\pi_k})^{m+\infty} v_{k-1}
        \\
        &= (\gamma P_{\pi_k})^m \sum_{j=0}^\infty((T_{\pi_k})^j v_{k-1} - (T_{\pi_k})^{j+1} v_{k-1})
        \\
        &= (\gamma P_{\pi_k})^m \sum_{j=0}^\infty (\gamma P_{\pi_k})^j (v_{k-1} -  T_{\pi_k} v_{k-1})
        \\
        &= (\gamma P_{\pi_k})^m (I-\gamma P_{\pi_k})^{-1} b_{k-1}^2.
    \end{align}
    
    To finish with, we prove the bound on the distance
    \begin{align}
        d_{k+1}^2 &= v_* - (T_{\pi_{k+1}})^m v_k
        \\
        &= \underbrace{T_{\pi_*} v_* - T_{\pi_*} v_k}_{ = \gamma P_{\pi_*} (v_* - v_k)}
        + \underbrace{T_{\pi_*} v_k - T_{\pi_{k+1}} v_k}_{\leq \epsilon'_{k+1} + \delta_k(\pi_*) \text{ by Lemma~\ref{lemma:key-lemma}}}
        + \underbrace{T_{\pi_{k+1}} v_k - (T_{\pi_{k+1}} v_k)^m}_{=\sum_{j=1}^{m-1} (\gamma P_{\pi_{k+1}})^j b_k^2}
        \\
        &=  \underbrace{\gamma P_{\pi_*} (v_* - v_k) + \gamma P_{\pi_*} \epsilon_k}_{= \gamma P_{\pi_*} (v_* - (v_k-\epsilon_k)) = \gamma P_{\pi_*} d_k^2}
        \underbrace{- \gamma P_{\pi_*} \epsilon_k + \epsilon'_{k+1}}_{=  y_k}
        + \delta_k(\pi_*)  + \sum_{j=1}^{m-1} (\gamma P_{\pi_{k+1}})^j b_k^2
        \\
        &= \gamma P_{\pi_*} d_k^2 + y_k + \sum_{j=1}^{m-1} (\gamma P_{\pi_{k+1}})^j b_k^2 + \delta_k(\pi_*).
    \end{align}
\end{proof}    

Now, we will show the component-wise bound on the regret $L_k$ of Thm.~\ref{th:component-wise-md-mpi}

\begin{proof}[Proof of Theorem~\ref{th:component-wise-md-mpi}]
    The proof is similar to the one of AMPI~\citep[Lemma~2]{scherrer2015approximate}, up to the additional term $\delta_k(\pi_*)$ and to the different bounded quantity. We will make use of the notation $\Gamma$, defined in Def.~\ref{def:Gamma}, and we will write $\Gamma_*$ if only the stochastic kernel induced by the optimal policy $\pi_*$ is involved. In other words, we write $\Gamma_*^j = (\gamma P_{\pi_*})^j$.
    
    From Lemma~\ref{lemma:bounds-bsd}, we have that 
    \begin{equation}
        d_{k+1}^h \leq \Gamma_* d_k^h + y_k + \sum_{j=1}^{m-1} \Gamma^j b_k^h + \delta_k(\pi_*).
    \end{equation}
    As the bound is the same for $h=1,2$, we remove the upperscript (and reintroduce it only when necessary, that is when going back to the core definition of these quantities).
    By direct induction, we get
    \begin{equation}
        d_k \leq \sum_{j=0}^{k-1} \Gamma_*^{k-1-j}\left(y_j + \sum_{l=1}^{m-1} \Gamma^l b_j + \delta_j(\pi_*)\right) + \Gamma^k d_0.
    \end{equation}
    Therefore, the loss $l_k$ can be bounded as (defining $\mathcal{L}_k$ at the same time)
    \begin{equation}
        l_k = d_k + s_k
        \leq \underbrace{\sum_{j=0}^{k-1} \Gamma_*^{k-1-j}(y_j + \sum_{l=1}^{m-1} \Gamma^l b_j) + \Gamma^k d_0 + s_k}_{=\mathcal{L}_k}  + \sum_{j=0}^{k-1} \Gamma_*^{k-1-j} \delta_j(\pi_*).
    \end{equation}
    The loss $\mathcal{L}_k$ is exactly of the same form as the one of AMPI, we'll take advantage of this later. From this bound on the loss $l_k$, we can bound the regret as:
    \begin{equation}
        L_k = \sum_{k=1}^K l_k \leq \sum_{k=1}^K \mathcal{L}_k + \sum_{k=1}^K \sum_{j=0}^{k-1} \Gamma_*^{k-1-j} \delta_j(\pi_*).
        \label{proof:component-md-mpi:regret1}
    \end{equation}
    We will first work on the last double sum. For this, we define $\Delta_j(\pi_*) = \sum_{k=0}^j \delta_k(\pi_*)$. We have
    \begin{align}
        \sum_{k=1}^K \sum_{j=0}^{k-1} \Gamma_*^{k-1-j} \delta_j(\pi_*) &= \sum_{k=0}^{K-1} \sum_{j=0}^{k} \Gamma_*^{k-j} \delta_j(\pi_*)
        \\
        &= \sum_{k=0}^{K-1} \sum_{j=0}^{k} \Gamma_*^j \delta_{k-j}(\pi_*)
        \\
        &= \sum_{j=0}^{K-1} \sum_{k=j}^{K-1} \Gamma_*^j \delta_{k-j}(\pi_*)
        \\
        &= \sum_{j=0}^{K-1} \Gamma_*^j \sum_{k=0}^{K-1-j} \delta_{k}(\pi_*) = \sum_{j=0}^{K-1} \Gamma_*^j \Delta_{K-1-j}(\pi_*).
    \end{align}
    For the last line, we used the fact that $\Gamma_*$ only involves the $P_{\pi_*}$ transition kernel, that does not depend on any iteration. Now, we can bound the term $\Delta_k(\pi_*)$, for any $k\geq 0$ as follows. Let write $R_{\Omega_{\pi_0}} = \|\sup_\pi D_\Omega(\pi||\pi_0)\|_\infty$, we have
    \begin{equation}
        \Delta_k(\pi_*) = \sum_{j=0}^k \delta_j(\pi_*) = \sum_{j=0}^k (D_\Omega(\pi_*||\pi_j) - D_\Omega(\pi_*||\pi_{j+1})) = D_\Omega(\pi_*||\pi_0) - D_\Omega(\pi_*||\pi_{k+1}) \leq D_\Omega(\pi_*||\pi_0) \leq R_{\Omega_{\pi_0}} \un.
    \end{equation}
    Given the definition of $\Gamma$, we have that $\Gamma^j \un = \gamma^j \un$, so
    \begin{equation}
        \sum_{k=1}^K \sum_{j=0}^{k-1} \Gamma_*^{k-1-j} \delta_j(\pi_*)
        \leq \sum_{j=0}^{K-1} \Gamma_*^j R_{\Omega_{\pi_0}} \un
        = \sum_{j=0}^{K-1} \gamma^j \un  R_{\Omega_{\pi_0}}
        = \frac{1-\gamma^K}{1-\gamma}   R_{\Omega_{\pi_0}} \un.
        \label{proof:component-md-mpi:telescopic}
    \end{equation}
    
    Next, we work on the term $\mathcal{L}_k$. As stated before, it is exactly the same as the one of the AMPI analysis, and the proof of~\citet[Lemma~2]{scherrer2015approximate} applies almost readily, we do not repeat it fully here. The only difference that appears and that induces a slight modification of the bound (on $\mathcal{L}_k)$) is how $b_0$ and $d_0$ are related, that will modify the $\eta_k$ term of the original proof. We will link $b_0$ and $d_0$ for both types of MD-MPI.
    
    For MD-MPI type~1 (with the natural convention that $\epsilon_0=0$), we have that
    \begin{equation}
        b^1_0 = v_0 - T_{\pi_1,\Omega_{\pi_0}} v_0 \text{ and } d^1_0 = v_* - (v_0 - \epsilon_0) = v_* - v_0.
    \end{equation}
    The Bellman residual can be written as
    \begin{align}
         b^1_0 &= v_0 - T_{\pi_1,\Omega_{\pi_0}} v_0
         \\
         &= \underbrace{v_0 - v_* + T_{\pi_*} v_* - T_{\pi_*} v_0}_{= (I-\gamma P_{\pi_*})(-d_0^1)}
         + \underbrace{T_{\pi_*} v_0 - T_{\pi_1,\Omega_{\pi_0}} v_0}_{\leq \epsilon'_1 + \delta_0(\pi_*) \text{ by Lemma~\ref{lemma:key-lemma}}}
         \\
         &\leq (I-\gamma P_{\pi_*})(-d_0^1) + \epsilon'_1 + D_\Omega(\pi_*||\pi_0)
         \\
         &\leq (I-\gamma P_{\pi_*})(-d_0^1) + \epsilon'_1 + R_{\Omega_{\pi_0}} \un,
    \end{align}
    where we used in the penultimate line the fact that $\delta_0(\pi_*) = D_\Omega(\pi_*||\pi_0) - D_\Omega(\pi_*||\pi_1) \leq D_\Omega(\pi_*||\pi_0)$ and in the last line the same bounding as before. So, the link between $b_0^1$ and $d_0^1$ is the same as for AMPI, up to the additional $R_{\Omega_{\pi_0}} \un$ term.
    
    For MD-MPI type~2, we have (with the same convention)
    \begin{equation}
        b_0^2 = v_0 - T_{\pi_1} v_0 \text{ and } d_0^2 = v_* - (v_0 - \epsilon_0) = v_* - v_0.
    \end{equation}
    Working on the Bellman residual
    \begin{align}
         b_0^2 &= v_0 - T_{\pi_1} v_0
         \\
         &= \underbrace{v_0 - v_* + T_{\pi_*} v_* - T_{\pi_*} v_0}_{= (I-\gamma P_{\pi_*})(-d_0^2)}
         + \underbrace{T_{\pi_*} v_0 - T_{\pi_1} v_0}_{\leq \epsilon'_1 + \delta_0(\pi_*) \text{ by Lemma~\ref{lemma:key-lemma}}}
         \\
         &\leq (I-\gamma P_{\pi_*})(-d_0^2) + \epsilon'_1 + D_\Omega(\pi_*||\pi_0) \leq (I-\gamma P_{\pi_*})(-d_0^2) + \epsilon'_1 + R_{\Omega_{\pi_0}} \un.
    \end{align}
    So, we have the same bound.
    
    Combining this with the part of the proof that does not change, and that we do not repeat, we get the following bound on $\mathcal{L}_k$:
    \begin{equation}
        \mathcal{L}_k \leq 2\sum_{i=1}^{k-1} \sum_{j=i}^\infty \Gamma^j |\epsilon_{k-i}|
        +  \sum_{i=0}^{k-1} \sum_{j=i}^\infty \Gamma^j |\epsilon'_{k-i}| + h(k) + \underbrace{\sum_{j=k}^\infty \Gamma^j \un R_{\Omega_{\pi_0}}}_{=\frac{\gamma^k}{1-\gamma}\un R_{\Omega_{\pi_0}}},
        \label{proof:component-md-mpi:loss-ampi}
    \end{equation}
    with $h(k)$ being defined as 
    \begin{equation}
        h(k) = 2\sum_{j=k}^\infty \Gamma^j |d_0| \text{ or } h(k) = 2\sum_{j=k}^\infty \Gamma^j |b_0|.
    \end{equation}
    
    Combining Eqs~\eqref{proof:component-md-mpi:telescopic} and~\eqref{proof:component-md-mpi:loss-ampi} into Eq.~\eqref{proof:component-md-mpi:regret1}, we can bound the regret:
    \begin{align}
        L_K &\leq \sum_{k=1}^K \mathcal{L}_k + \sum_{k=1}^K \sum_{j=0}^{k-1} \Gamma_*^{k-1-j} \delta_j(\pi_*)
        \\
        &\leq \sum_{k=1}^K \left(2\sum_{i=1}^{k-1} \sum_{j=i}^\infty \Gamma^j |\epsilon_{k-i}|
            +  \sum_{i=0}^{k-1} \sum_{j=i}^\infty \Gamma^j |\epsilon'_{k-i}| + h(k) + \frac{\gamma^k}{1-\gamma}\un R_{\Omega_{\pi_0}} \right)
        + \frac{1-\gamma^K}{1-\gamma}   R_{\Omega_{\pi_0}} \un
        \\
        &= 2\sum_{k=2}^K\sum_{i=1}^{k-1} \sum_{j=i}^\infty \Gamma^j |\epsilon_{k-i}|
        + \sum_{k=1}^K \sum_{i=0}^{k-1} \sum_{j=i}^\infty \Gamma^j |\epsilon'_{k-i}|
        + \sum_{k=1}^K h(k)
        + \frac{1-\gamma^K}{(1-\gamma)^2} R_{\Omega_{\pi_0}} \un.
    \end{align}
    This concludes the proof.
\end{proof}

To prove Cor.~\ref{cor:regret-md-mpi}, we will need a result from~\citet{scherrer2015approximate}, that we recall first.

\begin{lemma}[Lemma~6 of~\citet{scherrer2015approximate}]
    \label{lemma:concentrability}
    Let $\mathcal{I}$ and $(\mathcal{J}_i)_{i\in\mathcal{I}}$ be sets of non-negative integers, $\{\mathcal{I}_1, \dots, \mathcal{I}_n\}$ be a partition of $\mathcal{I}$, and $f$ and $(g_i)_{i\in\mathcal{I}}$ be functions satisfying
    \begin{equation}
        |f| \leq \sum_{i\in\mathcal{I}} \sum_{j\in\mathcal{J}_i} \Gamma^j |g_i| = \sum_{l=1}^n \sum_{i\in\mathcal{I}_l} \sum_{j\in\mathcal{J}_i} \Gamma^j |g_i|.
    \end{equation}
    Then, for all $p$, $q$ such that $\frac 1 p + \frac 1 q = 1$ and for all distributions $\rho$ and $\mu$, we have
    \begin{equation}
        \|f\|_{p,\rho} \leq \sum_{l=1}^n (\mathcal{C}_q(l))^{\frac 1 p} \sup_{i\in\mathcal{I}_l} \|g_i\|_{p q',\mu} \sum_{i\in\mathcal{I}_l}  \sum_{j\in\mathcal{J}_i} \gamma^j
    \end{equation}
    with the following concentrability coefficients,
    \begin{equation}
        \mathcal{C}_q(l) = \frac{\sum_{i\in\mathcal{I}_l}  \sum_{j\in\mathcal{J}_i} \gamma^j c_q(j)}{\sum_{i\in\mathcal{I}_l}  \sum_{j\in\mathcal{J}_i} \gamma^j}
        \text{, where } c_q(j) = \max_{\pi_1,\dots,\pi_j} \left\|\frac{\rho P_{\pi_1} P_{\pi_2}\dots P_{\pi_j}}{\mu}\right\|_{q,\mu}.
    \end{equation}
\end{lemma}

Now, we can prove the stated result.

\begin{proof}[Proof of Corollary~\ref{cor:regret-md-mpi}]
    The proof is an application of Lemma~\ref{lemma:concentrability} to Thm.~\ref{th:component-wise-md-mpi}. We define $\mathcal{I} = \{1,2,\dots, K^2 + K + 1\}$ and the associated trivial partition (that is, $\mathcal{I}_i = \{i\}$). For each $i\in\mathcal{I}$ we define
    \begin{align}
        g_i &=
        \begin{cases}
            2 \epsilon_{k-i'} &\text{ if } i = i'+\frac{(k-2)(k-1)}{2} \text{ with } 2\leq k \leq K \text{ and } 1\leq i' \leq k-1
            \\ 
            \epsilon'_{k-i'} &\text{ if } i = i'+\frac{k(k-1)}{2} + \frac{K(K-1)}{2} + 1 \text{ with } 1\leq k \leq K \text{ and } 0\leq i' \leq k-1
            \\
            2 d_0 \text{ or } 2 b_0 &\text{ if } i = K^2 + k \text{ with } 1\leq k \leq K
            \\
            \frac{1-\gamma^K}{(1-\gamma)^2} R_{\Omega_{\pi_0}} \un &\text{ if } i = K^2 + K + 1
        \end{cases}
        \\
        \text{and }
        \mathcal{J}_i &=
        \begin{cases}
            \{i',\dots\} &\text{ if } i = i'+\frac{(k-2)(k-1)}{2} \text{ with } 2\leq k \leq K \text{ and } 1\leq i' \leq k-1
            \\
            \{i',\dots\} &\text{ if } i = i'+\frac{k(k-1)}{2} + \frac{K(K-1)}{2} + 1 \text{ with } 1\leq k \leq K \text{ and } 0\leq i' \leq k-1
            \\
            \{k,\dots\} &\text{ if } i = K^2 + k \text{ with } 1\leq k \leq K
            \\
            \{0\} &\text{ if } i = K^2 + K + 1
        \end{cases}.
    \end{align}
    With this, Thm.~\ref{th:component-wise-md-mpi} rewrites as
    \begin{equation}
        L_K \leq \sum_{l=1}^{K^2 + K + 1} \sum_{i\in\mathcal{I}_l} \sum_{j \in \mathcal{J}_i} \Gamma^j |g_i|.
    \end{equation}
    The results follows by applying Lemma~\ref{lemma:concentrability} and using the fact that $\sum_{j\geq i} \gamma^j = \frac{\gamma^i}{1-\gamma}$
\end{proof}

The proof of Prop.~\ref{prop:loss-regret} is a basic application of the H\"older inequality.

\begin{proof}[Proof of Proposition~\ref{prop:loss-regret}]
    We write $\circ$ the Hadamard product. Recall that $l_k = v_* - v_{\pi_k} \geq 0$ and that $L_K = \sum_{k=1}^K l_k$. Notice that the sequence $v_{\pi_k}$ is not necessarily monotone, even without approximation error. On one side, we have that
    \begin{equation}
        \|L_K\|_{1,\rho} = \rho L_K = \sum_{k=1}^K \rho l_k \geq K \min_{1\leq k \leq K} \rho l_k = K \min_{1\leq k \leq K} \|l_k\|_{1,\rho}.
    \end{equation}
    On the other side, with $q$ such that $\frac 1 p + \frac 1 q = 1$ we have that
    \begin{equation}
        \rho L_K = \langle \rho^{\frac 1 p + \frac 1 q}, L_k \rangle
        = \langle \rho^{\frac 1 q}, \rho^{\frac 1 p} \circ L_K \rangle
        \leq \|\rho^{\frac 1 q}\|_q \|\rho^{\frac 1 p} \circ L_K\|_p
        = \|L_K\|_{p,\rho},
    \end{equation}
    where we used the H\"older inequality. Combining both equations provides the stated result. 
\end{proof}

Cor.~\ref{cor:regret-no-error} is a direct consequence of Cor.~\ref{cor:regret-md-mpi}.

\begin{proof}[Proof of Corollary~\ref{cor:regret-no-error}]
    Taking the limit of Cor.~\ref{cor:regret-md-mpi} as $p\rightarrow\infty$, when the errors are null, gives
    \begin{equation}
        \|L_K\|_\infty \leq 2\sum_{k=1}^K \frac{\gamma^k}{1-\gamma} \min(\|d_0\|_{\infty}, \|b_0\|_{\infty}) + \frac{1-\gamma^K}{(1-\gamma)^2} R_{\Omega_{\pi_0}} \leq 2 \gamma \frac{1 - \gamma^k}{(1-\gamma)^2} \|d_0\|_\infty + \frac{1-\gamma^K}{(1-\gamma)^2} R_{\Omega_{\pi_0}},
    \end{equation}
    where we used the fact that $\sum_{k=1}^K \gamma^k = \gamma \frac{1 - \gamma^K}{1-\gamma}$.
    The result follows by grouping terms and using $d_0 = v_* - v_0$.
\end{proof}

The proof of Cor.~\ref{cor:sum-errors} is mainly a manipulation of sums.

\begin{proof}[Proof of Corollary~\ref{cor:sum-errors}]
    From Cor.~\ref{cor:regret-md-mpi}, we have that
    \begin{equation}
        \|L_K\|_{p,\rho} \leq  2\sum_{k=2}^K\sum_{i=1}^{k-1} \frac{\gamma^i}{1-\gamma} (C_q^i)^{\frac{1}{p}} \|\epsilon_{k-i}\|_{pq',\mu}
        + \sum_{k=1}^K\sum_{i=0}^{k-1} \frac{\gamma^i}{1-\gamma} (C_q^i)^{\frac{1}{p}} \|\epsilon'_{k-i}\|_{pq',\mu}.
        + g(k) + \frac{1-\gamma^K}{(1-\gamma)^2} R_{\Omega_{\pi_0}}.
    \end{equation}
    We only need to work on the first two sums. To shorten the notations, write $\alpha_i =  \frac{\gamma^i}{1-\gamma} (C_q^i)^{\frac{1}{p}}$, $\beta_i = \|\epsilon_{k-i}\|_{pq',\mu}$ and $\beta'_i = \|\epsilon'_{k-i}\|_{pq',\mu}$. We have:
    \begin{align}
        \sum_{k=2}^K \sum_{i=1}^{k-1} \alpha_i \beta_{k-i} +  \sum_{k=1}^K \sum_{i=0}^{k-1} \alpha_i \beta'_{k-i}
        &= \sum_{k=1}^{K-1} \sum_{i=1}^{k} \alpha_i \beta_{k+1-i} +  \sum_{k=0}^{K-1} \sum_{i=0}^{k} \alpha_i \beta'_{k+1-i}
        \\
        &= \sum_{i=1}^{K-1} \alpha_i \sum_{k=i}^{K-1}  \beta_{k+1-i} +  \sum_{i=0}^{K-1} \alpha_i  \sum_{k=i}^{K-1} \beta'_{k+1-i}
        \\
        &= \sum_{i=1}^{K-1} \alpha_i \sum_{k=1}^{K-i}  \beta_{k} +  \sum_{i=0}^{K-1} \alpha_i  \sum_{k=1}^{K-i} \beta'_{k}.
    \end{align}
    The result follows by reinjecting this in Cor.~\ref{cor:regret-md-mpi} after having replaced $\alpha_i$, $\beta_i$ and $\beta'_i$.
\end{proof}

The proof of Cor.~\ref{cor:group-c} basically consists in the application of Lemma~\ref{lemma:concentrability} to a rewriting of Thm.~\ref{th:component-wise-md-mpi}.

\begin{proof}[Proof of Corollary~\ref{cor:group-c}]
    By Thm.~\ref{th:component-wise-md-mpi}, we have
    \begin{equation}
        L_K \leq 2\sum_{k=2}^K\sum_{i=1}^{k-1} \sum_{j=i}^\infty \Gamma^j |\epsilon_{k-i}|
        + \sum_{k=1}^K \sum_{i=0}^{k-1} \sum_{j=i}^\infty \Gamma^j |\epsilon'_{k-i}|
        + \sum_{k=1}^K h(k)
        + \frac{1-\gamma^K}{(1-\gamma)^2} R_{\Omega_{\pi_0}} \un.
        \label{proof:toto}
    \end{equation}
    We start by rewriting the two first sums. For the first one, we have
    \begin{align}
        \sum_{k=2}^K\sum_{i=1}^{k-1} \sum_{j=i}^\infty \Gamma^j |\epsilon_{k-i}|
        &= \sum_{k=2}^K\sum_{i=1}^{k-1} \sum_{j=k-i}^\infty \Gamma^j |\epsilon_{i}|
        \\
        &= \sum_{k=1}^{K-1}\sum_{i=1}^{k} \sum_{j=k+1-i}^\infty \Gamma^j |\epsilon_{i}|
        \\
        &= \sum_{i=1}^{K-1}\left(\sum_{k=i}^{K-1} \sum_{j=k+1-i}^\infty \Gamma^j\right) |\epsilon_{i}|
        \\
        &= \sum_{i=1}^{K-1}\left(\sum_{k=1}^{K-i} \sum_{j=k}^\infty \Gamma^j\right) |\epsilon_{i}|
        \\
        &= \sum_{i=1}^{K-1}\left(\sum_{k=1}^{i} \sum_{j=k}^\infty \Gamma^j\right) |\epsilon_{K-i}|.
    \end{align}
    Similarly, we have that
    \begin{equation}
        \sum_{k=1}^K \sum_{i=0}^{k-1} \sum_{j=i}^\infty \Gamma^j |\epsilon'_{k-i}|
        = \sum_{i=0}^{K-1} \left(\sum_{k=0}^i \sum_{j=k}^\infty \Gamma^j\right) |\epsilon'_{k-i}|.
    \end{equation}
    Thus, the bound on the loss can be writen as
    \begin{equation}
        L_K \leq 2\sum_{i=1}^{K-1}\left(\sum_{k=1}^{i} \sum_{j=k}^\infty \Gamma^j\right) |\epsilon_{K-i}|
        + \sum_{i=0}^{K-1} \left(\sum_{k=0}^i \sum_{j=k}^\infty \Gamma^j\right) |\epsilon'_{k-i}|
        + 2\left(\sum_{k=1}^K \sum_{j=k}^\infty \Gamma^j\right) (|d_0| \text{ or } |b_0|)
        + \frac{1-\gamma^K}{(1-\gamma)^2} R_{\Omega_{\pi_0}} \un.
    \end{equation}
    
    In order to apply Lemma~\ref{lemma:concentrability} to this bound, we consider $\mathcal{I} = \{1,2,\dots 2K +1\}$ and the associated trivial partition $\mathcal{I}_i = \{i\}$. For each $i\in\mathcal{I}$, we define:
    \begin{align}
        g_i &= \begin{cases}
            2 \epsilon_{K-i} &\text{ if } 1\leq i \leq K-1
            \\
            \epsilon'_{2K-i} &\text{ if } K\leq i \leq 2K-1
            \\
            2|d_0| \text{ or } 2|b_0| &\text{ if } i = 2K
            \\
            \frac{1-\gamma^K}{(1-\gamma)^2} R_{\Omega_{\pi_0}} \un  &\text{ if } i = 2K+1
        \end{cases}
        \\
        \text{and }
        \mathcal{J}_i &= \begin{cases}
            \bigcup_{k=1}^i \{k,k+1,\dots\} &\text{ if } 1\leq i \leq K-1
            \\
            \bigcup_{k=0}^{i-K} \{k,k+1,\dots\} &\text{ if } K\leq i \leq 2K-1
            \\
            \bigcup_{k=1}^{K} \{k,k+1,\dots\}  &\text{ if } i = 2K
            \\
            \{0\}  &\text{ if } i = 2K+1
        \end{cases}.
    \end{align}
    With this, Eq.~\eqref{proof:toto} rewrites as
    \begin{equation}
        L_K \leq \sum_{l=1}^{2K + 1} \sum_{i\in\mathcal{I}_l} \sum_{j \in \mathcal{J}_i} \Gamma^j |g_i|.
    \end{equation}
    With the $c_q$ term defined in Lemma~\ref{lemma:concentrability}, using the fact that $\sum_{i=l}^{k-1} \sum_{j=i}^\infty \gamma^j = \frac{\gamma^l - \gamma^k}{(1-\gamma)^2}$, as well as the following concentrability coefficient,
    \begin{equation}
        C_q^{l,k} = \frac{(1-\gamma)^2}{\gamma^l - \gamma^k} \sum_{i=l}^{k-1}\sum_{j=i}^\infty c_q(j),
    \end{equation}
    we get the following bound on the regret:
    \begin{align}
        \|L_K\|_{p,\rho} &\leq 2 \sum_{i=1}^{K-1} \frac{\gamma -\gamma^{i+1}}{(1-\gamma)^2} (C_q^{1,i+1})^{\frac 1 p} \|\epsilon_{k-i}\|_{pq',\mu}
        +
        \sum_{i=0}^{K-1} \frac{1 -\gamma^{i+1}}{(1-\gamma)^2} (C_q^{0,i+1})^{\frac 1 p } \|\epsilon'_{k-i}\|_{pq',\mu}
        \\
        &\quad +
        \frac{\gamma -\gamma^{K+1}}{(1-\gamma)^2} (C_q^{1,K+1})^{\frac 1 p}\min(\|d_0\|_{pq',\mu},\|b_0\|_{pq',\mu})
        +
        \frac{1-\gamma^K}{(1-\gamma)^2} R_{\Omega_{\pi_0}}.
    \end{align}
\end{proof}

\section{Perspectives on regularized MDPs}
\label{appx:discussion-perspectives}

Here, we discuss in more details the perspectives briefly mentioned in Sec.~\ref{sec:conclusion}.

\subsection{Dynamic programming and optimization}

We have shown how MPI regularized by a Bregman divergence is related to Mirror Descent. The computation of the regularized greedy policy is similar to a mirror descent step, where the $q$-function plays the role of the negative subgradient. In this sense, the policy lives in the primal space while the $q$-function lives in the dual space. It would be interesting to take inspiration from proximal convex optimization to derive new dynamic programming approaches, for example based on Dual Averaging~\citep{nesterov2009primal} or Mirror Prox~\citep{nemirovski2004prox}.

For example, consider the case of a fixed regularizer. We have seen that it leads to a different solution than the original one (see Thm.~\ref{th:reg-opt-in-original-mdp}). Usually, one considers a scaled negative entropy as such a regularizer\footnote{This is the case for SAC or soft Q-learning, for example. Sometimes, it is the reward that is scaled, but both are equivalent.}, $\Omega(\pi(\cdot|s)) = \alpha \sum_a \pi(a|s)\ln\pi(a|s)$. The choice of this parameter is important practically and problem-dependent: too high and the solution of the regularized problem will be very different from the original one, too low and the algorithm will not benefit from the regularization. A natural idea is to vary the weight of the regularizer over iterations~\citep{peters2010relative,abdolmaleki2018relative,haarnoja2018softb}, much like a learning rate in a gradient descent approach.

More formally, in our framework, write $\Omega_k = \alpha_k \Omega$ and consider the following weighted reg-MPI scheme,
\begin{equation}
    \begin{cases}
        \pi_{k+1} = \gc^{\epsilon'_{k+1}}_{\Omega_k}(v_k)
        \\
        v_{k+1} = (T_{\pi_{k+1},\Omega_k})^m v_k + \epsilon_{k+1}
    \end{cases},
\end{equation}
with $\gc^{\epsilon'_{k+1}}_{\Omega_k}(v_k)$ as defined in Sec.~\ref{subsec:analysis-regMPI}: for any policy $\pi$, $T_{\pi,\Omega_k} v_k \leq T_{\pi_{k+1},\Omega_k} v_k + \epsilon'_{k+1}$. By applying the proof techniques developed previously, one can obtain easily the following result.

\begin{theorem}
    \label{th:weighted-regMPI}
    Define $R_{\Omega}= \|\sup_{\pi} \Omega(\pi)\|_\infty$. Assume that the series $(\alpha_k)_{k\geq 0}$ is positive and decreasing, and that the regularizer $\Omega$ is positive (without loss of generality). After $K$ iterations of the preceding weighted reg-MPI scheme, the regret satisfies
    \begin{align}
        L_K \leq 2\sum_{k=2}^K\sum_{i=1}^{k-1} \sum_{j=i}^\infty \Gamma^j |\epsilon_{k-i}|
    + \sum_{k=1}^K \sum_{i=0}^{k-1} \sum_{j=i}^\infty \Gamma^j |\epsilon'_{k-i}|
    + \sum_{k=1}^K h(k)
    + \frac{1-\gamma^K}{(1-\gamma)^2} R_{\Omega} \sum_{k=0}^{K-1}\alpha_k \un.
    \end{align}
    with $h(k) = 2\sum_{j=k}^\infty \Gamma^j |d_0|$ or $h(k) = 2\sum_{j=k}^\infty \Gamma^j |b_0|$.
\end{theorem}
\begin{proof}
    The proof is similar to the one of MD-MPI, as it bounds the regret, but also simpler as it does not require Lemma~\ref{lemma:key-lemma}. The principle is still to bound the distance $d_k$ and the shift $s_k$, that both require bounding the Bellman residual $b_k$. From this, one can bound the loss $l_k = d_k + s_k$, and then the regret $L_K = \sum_{k=1}^K l_k$. We only specify what changes compared to the previous proofs.
    
    Bounding the shift $s_k$ is done as before, and one get the same bound:
    \begin{equation}
        s_k = (T_{\pi_k,\Omega_{k-1}})^m v_{k-1} - v_{\pi_k} \leq (\gamma P_{\pi_k})^m (I-\gamma P_{\pi_k}) b_{k-1}.
    \end{equation}
    For the distance $d_k$, we have
    \begin{align}
        d_{k+1} &= v_* - (T_{\pi_{k+1},\Omega_{k}})^m v_{k}
        \\
        &= \underbrace{T_{\pi_*} v_* - T_{\pi_*} v_k}_{ = \gamma P_{\pi_*}(v_* - v_k)}
        + \underbrace{T_{\pi_*} v_k - T_{\pi_*,\Omega_k} v_k}_{= \Omega_k(\pi_*) \leq \alpha_k R_\Omega \un}
        + \underbrace{ T_{\pi_*,\Omega_k} v_k -  T_{\pi_{k+1},\Omega_k} v_k}_{\leq \epsilon'_{k+1}}
        + T_{\pi_{k+1},\Omega_k} v_k - (T_{\pi_{k+1},\Omega_{k}})^m v_{k}.
    \end{align}
    The rest of the bounding is similar to MD-MPI, and we get
    \begin{equation}
        d_{k+1} \leq \gamma P_{\pi_*} d_k + y_k + \alpha_k R_\Omega\un + \sum_{j=1}^{m-1} (\gamma P_{\pi_{k+1}})^j b_k.
    \end{equation}
    So, this is similar to the bound of MD-MPI, with the term $\alpha_k R_\Omega\un$ replacing the term $\delta_k(\pi_*)$. For the Bellman residual, we have
    \begin{equation}
        b_k = v_k - T_{\pi_{k+1},\Omega_k} v_k
        = \underbrace{v_k - T_{\pi_k,\Omega_k} v_k}_{\leq v_k -  T_{\pi_{k},\Omega_{k-1}} v_k}
        + \underbrace{T_{\pi_k,\Omega_k} v_k -  T_{\pi_{k+1},\Omega_k} v_k}_{\epsilon'_{k+1}}
    \end{equation}
    where we used the facts that $\alpha_k \leq \alpha_{k-1}$ and that $\Omega\geq 0$ for bounding the first term:
    \begin{equation}
        - T_{\pi_k,\Omega_k} v_k
        = - T_{\pi_k} v_k + \alpha_k \Omega(\pi_k)
        \leq - T_{\pi_k} v_k + \alpha_{k-1} \Omega(\pi_k) = - T_{\pi_k,\Omega_{k-1}} v_k.
    \end{equation}
    The rest of the bounding is as before and gives $b_k \leq (\gamma P_{\pi_k})^m b_{k-1} + x_k$. From these bounds, one can bound $L_k$ as previously.
\end{proof}

From Thm.~\ref{th:weighted-regMPI}, we can obtain an $\ell_p$-bound on the regret, and from this the rate of convergence of the average regret. For MD-MPI, the rate of convergence was in $\mathcal{O}(\frac 1 K)$. Here, it depends on the weighting of the regularizer. For example, if $\alpha_k$ is in $\mathcal{O}(\frac{1}{k})$, then the average regret will be in $\mathcal{O}(\frac{\ln K}{K})$. If $\alpha_k$ is in $\mathcal{O}(\frac{1}{\sqrt k})$, then the average regret will be in $\mathcal{O}(\frac{1}{\sqrt{K}})$. This illustrates the kind of things that can be done with the proposed framework of regularized MDPs.

\subsection{Temporal consistency equations}

Thanks to $\Omega$, the regularized greedy policies are unique, and thus is the regularized optimal policy. The pair of optimal policy and optimal value function can be characterized as follows.

\begin{cor}
    The optimal policy and optimal value function in a regularized MDP are the unique functions satisfying
    \begin{equation}
         \forall s\in\s, \quad \begin{cases}
        v_{*,\Omega}(s) = \Omega^*\left(r(s,\cdot) + \gamma \E_{s'|s,\cdot}[v_{*,\Omega}(s')]\right)
        \\
        \pi_{*,\Omega}(\cdot|s) = \nabla \Omega^*\left(r(s,\cdot) + \gamma \E_{s'|s,\cdot}[v_{*,\Omega}(s')]\right)
    \end{cases}.
    \end{equation}
\end{cor}
\begin{proof}
    This is a direct consequence of Thm.~\ref{th:opt-reg-pol}
\end{proof}

This provides a general way to estimate the optimal policy. For example, if $\Omega$ is the negative entropy, this set of equations simplifies to
\begin{equation}
    \forall (s,a)\in\s\times\A \quad
    v_{*,\Omega}(s) = r(s,a) + \gamma \E_{s'|s,a}[v_{*,\Omega}(s')] - \ln \pi_{*,\Omega}(a|s).
\end{equation}
This has been used by~\citet{nachum2017bridging} or~\citet{dai2018sbeed}, where it is called ``temporal consistency equation'', to estimate the optimal value-policy pair by minimizing the related residual,
\begin{equation}
    \E_{s,a}\left[\left(r(s,a) + \gamma \E_{s'|s,a}[v_{*,\Omega}(s')] -  \ln \pi_{*,\Omega}(a|s) - v_{*,\Omega}(s)\right)^2\right].
\end{equation}

This idea has also been extended to Tsallis entropy~\citep{nachum2018path}. In this case, the set of equations does not simplify as nicely as with the Shannon entropy. Instead, the approach consists in considering the Lagrangian derived from the Legendre-Fenchel transform (and the resulting temporal consistency equation involves Lagrange multipliers, that have to be learnt too). This idea could be extended to other regularizers. One could also replace the regularizer $\Omega$ by a Bregman divergence, and estimate the optimal policy by solving a sequence of temporal consistency equations.  

\subsection{Regularized policy gradient}

Policy search approaches often combine policy gradient with an entropic regularization, typically to prevent the policy from becoming too quickly deterministic~\cite{williams1992simple,mnih2016asynchronous}. The policy gradient theorem~\cite{sutton2000policy} can easily be extended to the proposed framework.

Let $\nu$ be a (user-defined) state distribution, the classical policy search approach consists in maximizing $J(\pi) = \E_{s\sim\nu}[v_\pi(s)] = \nu v_\pi$. This principle can easily be extended to regularized MDPs, by maximizing
\begin{equation}
    J_\Omega(\pi) = \nu v_{\pi,\Omega}.
\end{equation}
Write $d_{\nu,\pi}$ the $\gamma$-weighted occupancy measure induced by the policy $\pi$ when the initial state is sampled from $\nu$, defined as $d_{\nu,\pi} = (1-\gamma)\nu (I-\gamma P_\pi)^{-1} \in \Delta_\s$. Slightly abusing notations, we'll also write $d_{\nu,\pi}(s,a) = d_{\nu,\pi}(s)\pi(a|s)$.

\begin{theorem}[Policy gradient for regularized MDPs]
    The gradient of $J_\Omega$ is
    \begin{equation}
    \nabla J_\Omega(\pi) = 
        \frac{1}{1-\gamma} \E_{s,a\sim d_{\nu,\pi}}\left[\left(q_{\pi,\Omega}(s,a) - \frac{\partial \Omega(\pi(.|s))}{\partial \pi(a|s)}\right)\nabla \ln \pi(a|s)\right].
    \end{equation}
\end{theorem}
\begin{proof}
    We have that
    \begin{equation}
        \nabla J_\Omega(\pi) = \sum_{s\in\s} \nu(s) \nabla v_{\pi,\Omega}(s).
    \end{equation}
    We have to study the gradient of the value function. For this, the nabla-log trick is useful: $\nabla \pi = \pi \nabla \ln \pi$.
    \begin{align}
        \nabla v_{\pi,\Omega}(s) &= \nabla\left(\sum_{a\in\A} \pi(a|s) (r(s,a) + \gamma\E_{s'|s,a}[v_{\pi,\Omega}(s')]) - \Omega(\pi(.|s))\right)
        \\
        &= \sum_{a\in\A} \left(\nabla \pi(a|s) q_{\pi,\Omega}(s,a) + \pi(a|s) \gamma\E_{s'|s,a}[\nabla v_{\pi,\Omega}(s')] \right) - \nabla \Omega (\pi(.|s))
        \\
        &= \sum_{a\in\A} \pi(a|s) \left(q_{\pi,\Omega}(s,a)\nabla\ln\pi(a|s) - \nabla \Omega (\pi(.|s)) + \gamma \E_{s'|s,a}[\nabla v_{\pi,\Omega}(s')]\right).
    \end{align}
    So, the components of $\nabla v_{\pi,\Omega}$ are the (unregularized) value functions corresponding to the rewards being the components of $q_{\pi,\Omega}(s,a)\nabla\ln\pi(a|s) - \nabla \Omega (\pi(.|s))$. Consequently,
    \begin{equation}
        J_\Omega(\pi) = \frac{1}{1-\gamma} \sum_{s\in\s} d_{\nu,\pi}(s)\sum_{a\in\A} \pi(a|s)(q_{\pi,\Omega}(s,a)\nabla\ln\pi(a|s) - \nabla \Omega (\pi(.|s))).
    \end{equation}
    Using the chain-rule, we have that
    \begin{equation}
        \nabla \Omega (\pi(.|s)) = \sum_{a\in\A} \frac{\partial \Omega(\pi(.|s))}{\partial \pi(a|s)} \nabla\pi(a|s)
        = \sum_{a\in\A} \frac{\partial \Omega(\pi(.|s))}{\partial \pi(a|s)} \pi(a|s) \nabla\ln\pi(a|s).
    \end{equation}
    Injecting this in the previous result concludes the proof.
\end{proof}

Even in the entropic case, it might be not exactly the same as the usual regularized policy gradient, notably because our result involves the regularized $q$-function. Again, the regularizer $\Omega$ could be replaced by a Bregman divergence, to ultimately estimate the optimal policy of the original MDP. It would be interesting to compare empirically the different resulting policy gradients approaches, we left this for future work.

\subsection{Regularized inverse reinforcement learning}

Inverse reinforcement learning (IRL) consists in finding a reward function that explains the behavior of an expert which is assumed to act optimally. It is often said that it is an ill-posed problem. The classical example is the null reward function that explains any behavior (as all policies are optimal in this case).

We argue that in this regularized framework, the problem is not ill-posed, because the optimal policy is unique, thanks to the regularization. For example, if one consider the negative entropy as the regularizer, with a null reward, the optimal policy will be that of maximum entropy, so the uniform policy, and it is unique.

Notice that if for a reward, the associated regularized optimal policy is unique, the converse is not true. For example, the uniform policy is optimal for any constant reward. More generally, reward shaping~\citep{ng1999policy} still holds true for regularized MDPs (this being thanks to the results of Sec.~\ref{sec:reg-mpds}, again).

This being said, assume that the model (dynamic, discount factor, regularizer) and that the optimal regularized policy $\pi_{*,\Omega}$ are known. It is possible to retrieve a reward function such that $\pi_{*,\Omega}$ is the unique optimal policy.

\begin{prop}
    Let $\hat{q}\in\R^{\s\times\A}$ be any function satisfying 
    \begin{equation}
        \forall s\in\s, \quad \pi_{*,\Omega}(\cdot|s) = \nabla\Omega^*(\hat{q}(s,\cdot)),
    \end{equation}
    then the reward $\hat{r}(s,a)$ defined as
    \begin{align}
        \forall (s,a)\in\s\times\A, \quad \hat{r}(s,a) &= \hat{q}(s,a) - \gamma \E_{s'|s,a}[\Omega^*(\hat{q}(s,\cdot))]
         \\
         &= \hat{q}(s,a) - \gamma \E_{s'|s,a}[\E_{a'\sim\pi_{*,\Omega}(\cdot|s')}[\hat{q}(s,a)] - \Omega(\pi_{*,\Omega}(\cdot|s))]
    \end{align}
    has $\pi_{*,\Omega}$ as the unique corresponding optimal policy.
\end{prop}
\begin{proof}
    First, recall (see Prop.~\ref{prop:convex-conjugate}) that if for any $q$, there is a unique regularized greedy policy, the converse is not true (simply by the fact that for any $v\in\R^{\s}$, we have $\nabla\Omega^*(q(s,\cdot)+v(s)) = \nabla\Omega^*(q(s,\cdot))$). By assumption and by uniqueness of regularized greediness, $\pi_{*,\Omega}$ is the unique regularized policy corresponding to $\hat{q}$. Then, with the above defined reward function, $\hat{q}$ is unique the solution of the regularized Bellman optimality equation. This shows the stated result.
\end{proof}

If this result shows that IRL is well-defined in a regularized framework, it is not very practical (for example, in the entropic case, it tells that $\hat{r}(s,a) = \ln \pi_{*,\Omega}(s,a)$ is such a reward function). Yet, we think that the proposed general framework could lead to more practical algorithm.

For example, many IRL algorithms are based on the maximum-entropy principle, eg.~\cite{ziebart2008maximum,finn2016guided,fu2017learning}. This maximum-entropy IRL framework can be linked to probabilistic inference, that can itself be shown to be equivalent, in some specific cases (deterministic dynamics), to entropy-regularized reinforcement learning~\cite{levine2018reinforcement}. We think this to be an interesting connection, and maybe that our proposed regularized framework could allow to generalize or analyze some of these approaches.

\subsection{Regularized zero-sum Markov games}

A zero-sum Markov game can be seen as a generalization of MDPs. It is a tuple $\{\s,\A^1,\A^2,P,r,\gamma\}$ with $\s$ the state space common to both players, $\A^j$ the action space of player $j$, $P\in\Delta_{\s}^{\s\times\A_1\times\A_2}$ the transition kernel ($P(s'|s,a^1,a^2)$ is the probability of transiting to state $s'$ when player~1 played $a^1$ and player~2 played $a^2$ in $s$), $r\in\R^{\s\times\A^1\times \A^2}$ the reward function of both players (one tries to maximize it, the other one to minimize it) and $\gamma$ the discount factor. We write $\mu\in\Delta_{\A^1}^\s$ a policy of the maximizer, and $\nu\in\Delta_{\A^2}^\s$ a policy of the minimizer.

As in the case of classical MDPs, everything can be constructed from an evaluation operator, defined as
\begin{equation}
		[T_{\mu,\nu} v](s) = \sum_{a^1} \sum_{a^2} \mu(a^1|s) \nu(a^2|s)\left(r(s,a^1,a^2) + \gamma \sum_{s'} p (s'|s,a^1,a^2) v(s')\right),
\end{equation}
of fixed point $v_{\mu,\nu}$.
From this, the following operators are defined:
\begin{align}
    T_\mu v &= \min_\nu T_{\mu,\nu} v
    &
    T_{\nu} v &= \max_\mu T_{\mu,\nu} v
    \\
    T v &= \max_\mu T_\mu v
    &
    \hat{T}_v &= \min_\nu T_{\nu} v.
\end{align}
We also define the greedy operator as $\mu\in\g(v) \Leftrightarrow T v = T_\mu v = \min_\nu T_{\mu,\nu} v$.
Thanks to the Von Neumann's minimax theorem~\cite{morgenstern1953theory}, we have that
\begin{equation}
    \hat{T} v = T v
\end{equation}
and the optimal value function satisfies
\begin{equation}
    v_* = \min_\nu\max_\mu v_{\mu,\nu} = \max_\mu\min_\nu v_{\mu,\nu}.
\end{equation}
The modified policy iteration for this kind of games is
\begin{equation}
    \begin{cases}
        \mu_{k+1} \in \g(v_k)
        \\
        v_{k+1} = T_{\mu_{k+1}}^m v_k
    \end{cases}.\label{eq:mpi-game}
\end{equation}

As in MDPs, we can regularize the evaluation operator, and construct regularized zero-sum Markov games from this. Let $\Omega_1$ and $\Omega_2$ be two strongly convex reguralizers on $\Delta_{\A_1}$ and $\Delta_{\A_2}$ , and define the regularized evaluation operator as
\begin{equation}
    [T_{\mu,\nu,\Omega} v](s) = [T_{\mu,\nu} v](s) - \Omega_1(\mu(.|s))  + \Omega_2(\nu(.|s)).
\end{equation}
From this, we can construct a theory of regularized zero-sum Markov games as it was done for MDPs. The Von Neumann's minimax theorem does not only hold for affine operators, but for convex-concave operators, so we're fine.

Notably, the unregularized error propagation analysis of~\eqref{eq:mpi-game} by~\citet{perolat2015approximate} could be easily adapted to the regularized case (much like how the analysis of AMPI directly led to the analysis of Sec.~\ref{subsec:analysis-regMPI}, thanks to the results of Sec.~\ref{sec:reg-mpds}). We left its extension to regularization with a Bregman divergence as future work.

\end{document}